\documentclass{article}

\PassOptionsToPackage{numbers, compress}{natbib}
 \usepackage[preprint]{neurips_2025}

\usepackage[utf8]{inputenc} 
\usepackage[T1]{fontenc}    
\usepackage[hidelinks]{hyperref}
\usepackage{url}            
\usepackage{booktabs}       
\usepackage{amsfonts}       
\usepackage{nicefrac}       
\usepackage{microtype}      
\usepackage{xcolor}         
\usepackage{graphicx}
\usepackage{amsmath}
\usepackage{algorithm}
\usepackage{algorithmicx}
\usepackage{algpseudocode}
\usepackage{multirow}
\usepackage{listings}
\usepackage[most]{tcolorbox} 
\usepackage{authblk}
\makeatletter
\renewcommand\AB@affilsepx{\quad\protect\Affilfont} 
\makeatother

\title{Global Commander and Local Operative: A Dual-Agent Framework for Scene Navigation}

\author[1]{\textbf{Kaiming Jin}\textsuperscript{*}}
\author[2]{\textbf{Yuefan Wu}}
\author[3]{\textbf{Shengqiong Wu}\textsuperscript{\dag}} 
\author[1]{\textbf{Bobo Li}}
\author[1]{\textbf{Shuicheng Yan}}
\author[1]{\textbf{Tat-Seng Chua}}

\affil[1]{National University of Singapore}
\affil[2]{Simon Fraser University}
\affil[3]{University of Oxford}

\begin{document}

\maketitle
\renewcommand{\thefootnote}{\fnsymbol{footnote}} 
\insert\footins{\noindent\footnotesize \kern2pt \textsuperscript{*}kaiming.jin@u.nus.edu \\ \textsuperscript{\dag}Corresponding author.}
\makeatletter
\def\@makefnmark{}
\makeatother

\begin{abstract}
Vision-and-Language Scene navigation is a fundamental capability for embodied human–AI collaboration, requiring agents to follow natural language instructions to execute coherent action sequences in complex environments.
Existing approaches either rely on multiple agents, incurring high coordination and resource costs, or adopt a single-agent paradigm, which overloads the agent with both global planning and local perception, often leading to degraded reasoning and instruction drift in long-horizon settings.
To address these issues, we introduce \textbf{DACo}, a planning–grounding decoupled architecture that disentangles global deliberation from local grounding.
Concretely, it employs a \textbf{Global Commander} for high-level strategic planning and a \textbf{Local Operative} for egocentric observing and fine-grained execution. 
By disentangling global reasoning from local action, DACo alleviates cognitive overload and improves long-horizon stability. The framework further integrates dynamic subgoal planning and adaptive replanning to enable structured and resilient navigation.
Extensive evaluations on R2R, REVERIE, and R4R demonstrate that DACo achieves 4.9\%, 6.5\%, 5.4\% absolute improvements over the best-performing baselines in zero-shot settings, and generalizes effectively across both closed-source (e.g., GPT-4o) and open-source (e.g., Qwen-VL Series) backbones. 
DACo provides a principled and extensible paradigm for robust long-horizon navigation. 
The code is available at: 
\href{https://github.com/ChocoWu/DACo}{DACo.io.}
\end{abstract}

\section{Introduction}

Building intelligent agents that can interpret human intent, perceive complex environments, and act coherently is central to human–machine collaboration~\citep{marsili2025visual,ren2023human}. 
Such capabilities underpin a wide spectrum of applications, including embodied assistants~\citep{zheng2024towards,huang2024vinci}, collaborative robots~\citep{sherwani2020collaborative,werby2024hierarchical}, and mixed-reality systems~\citep{schwede2015holor,bohus2024sigma}. 
Among these competencies, vision-language navigation (VLN)~\cite{mattersim,Matterport3D,reverie,R2R} serves as a canonical testbed, requiring agents to follow natural language instructions to navigate complex indoor environments and reach designated targets.
Considering its practicality and flexibility in facilitating real-world human–robot interaction, scene navigation has garnered significant research attention within the human–machine interaction community.

\begin{figure}[h]
    \centering
    \includegraphics[width=\textwidth]{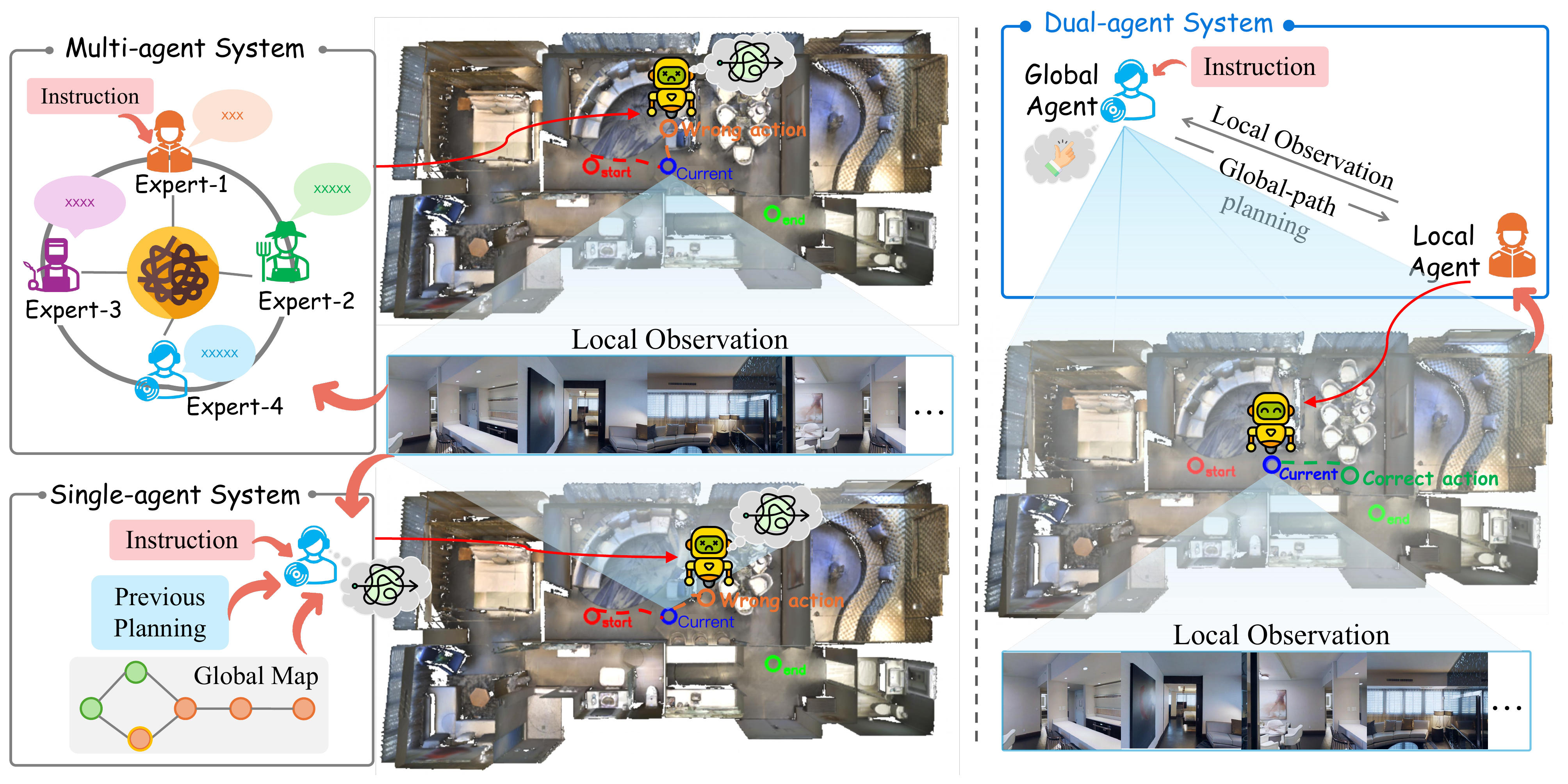}
    \vspace{-2mm}
   \caption{Comparison of different agentic systems for scene navigation. Multi-agent systems rely on multiple experts, resulting in high coordination overhead and resource costs. Single-agent systems must handle both global planning and local perception, overloading the decision process. In contrast, our dual-agent framework assigns clear and complementary roles to global and local agents, simplifying system design and enabling more robust navigation reasoning.}
    \label{fig:intro}
\end{figure}

Early explorations in scene navigation primarily focused on constructing task-specific pipelines through the design of specialized architectures~\citep{wang2019reinforced, ma2019self, deng2020evolving, qi2020object}, advanced learning paradigms~\citep{zhu2020vision,li2019robust}, and data augmentation techniques~\citep{fried2018speaker,fu2020counterfactual,gu2022vision}.
With the advent of large pre-trained models~\citep{he2016deep,neil2020transformers}, subsequent works~\citep{chen2021hamt,chen2022think,hong2021vln,qi2021road} have leveraged these models through domain-specific fine-tuning to enhance cross-modal alignment and enable more effective end-to-end learning, thereby significantly reducing the need for hand engineering.
However, the limited scale of VLN datasets compared to large-scale pre-training corpora restricts the acquisition of robust navigational reasoning, particularly in unseen environments.

The recent emergence of large vision-language models (LVLMs)~\citep{yang2025qwen3,liu2023llava,bai2025qwen25vl}, which exhibit strong commonsense reasoning and cognitive abilities, has opened new opportunities for building more flexible and generalizable navigation systems.
Recent works~\citep{zu2024language,marsili2025visual,lin2025evolvenav,zhou2024navgpt2,perincherry2025visual} have explored using open-source LVLMs as the backbone for scene navigation through additional fine-tuning on in-domain datasets. 
This strategy leverages the rich knowledge embedded in LVLMs to enhance navigation tasks, yielding impressive performance improvements.
However, fine-tuning large models requires substantial GPU resources and high-quality, domain-specific data, which limits their scalability.
To circumvent these limitations, many studies have turned to training-free paradigms, leveraging powerful closed-source LVLMs such as GPT series~\citep{OpenAI2024GPT4o} to construct agent systems for navigation reasoning.
As illustrated in Fig.~\ref{fig:intro}, existing designs primarily follow two paradigms.
Multi-agent systems (Fig.~\ref{fig:intro}, top-left) ~\citep{navgpt,discussnav,wang2025dreamnav,wei2025unseen, canav} employ multiple agents or expert modules to maximize reasoning capabilities; however, this design incurs significant GPU and token costs and introduces considerable challenges in expert coordination. 
In contrast, single-agent systems (Fig.~\ref{fig:intro}, bottom-left) ~\citep{mapgpt, liu2025msnav} rely on a single-agent paradigm, where the agent integrates topological maps for global context with local observations to achieve a comprehensive understanding.
Although computationally simpler, this design entangles heterogeneous information streams and forces a single model to simultaneously perform long-horizon planning and fine-grained action grounding, often leading to cognitive overload, degraded spatial reasoning, and instruction drift, particularly in long trajectories. 
These limitations reveal a structural tension in LVLM-based navigation: strong reasoning requires specialization, yet excessive modularity introduces coordination complexity. 
The key is therefore not increasing capacity, but disentangling fundamentally different competencies with minimal architectural overhead.

To this end, we propose \textbf{DACo}, a minimal \underline{D}ual-\underline{A}gent \underline{Co}llaboration architecture grounded in cognitive decomposition, as depicted in Fig.~\ref{fig:intro} (right). 
Rather than treating navigation as a monolithic reasoning problem, DACo factorizes it into two complementary competencies: a \emph{Global Commander}, responsible for top-down strategic planning over global context, and a \emph{Local Operative}, dedicated to egocentric grounding and fine-grained action execution. 
This structured separation disentangles global deliberation from local perception, reducing contextual interference and improving long-horizon stability.
Building upon this decomposition, DACo introduces dynamic subgoal planning to provide intermediate guidance and an adaptive replanning mechanism to correct trajectory deviations. 
Together, these components enable structured, interpretable, and resilient navigation.

We evaluate DACo on three widely used benchmarks, i.e., R2R~\citep{R2R}, REVERIE~\citep{reverie}, and R4R~\citep{R4R}, representing diverse evaluative focuses and task complexities.
DACo achieves substantial performance improvements in zero-shot settings, outperforming the best-performing baselines by 4.9\% on R2R, 6.5\% on REVERIE, and 5.4\% on R4R.
Moreover, DACo demonstrates consistent gains across both closed-source (e.g., GPT-4o~\citep{OpenAI2024GPT4o}) and open-source (e.g., Qwen2.5-VL-32B\citep{bai2025qwen25vl} and Qwen3-VL-8B~\citep{bai2025qwen3vltechnicalreport}) backbones, and shows particular advantages in long-horizon navigation tasks.
In summary, our contributions are threefold:
\begin{itemize}

\item We propose \textbf{DACo}, a novel minimal role-specialized dual-agent architecture that structurally decomposes global planning and local execution for LVLM-based navigation.

\item We introduce dynamic subgoal planning and adaptive replanning mechanisms that enhance stability and interpretability in long-horizon navigation.

\item We conduct extensive evaluations across multiple benchmarks and backbones, demonstrating consistent and significant zero-shot improvements.
\end{itemize}

\section{Related Work}

\subsection{Vision-Language Navigation}
Vision-Language navigation (VLN), which aims to build embodied agents capable of following natural language instructions to achieve a goal in virtual 3D environments, has garnered extensive interest~\citep{reverie,R2R}.
A variety of benchmarks have been introduced to assess agents' abilities in instruction following and spatial reasoning.
Early approaches~\citep{wang2019reinforced,ma2019self,deng2020evolving,zhu2020vision,tan2019learning} primarily focused on learning-based methods trained on domain-specific datasets. 
With the rise of pre-trained models~\citep{he2016deep,neil2020transformers}, researchers have begun leveraging large-scale multimodal representations to improve instruction grounding and action prediction~\citep{zheng2024towards,chen2022think,hong2021vln,wang2023scaling,qiao2022hop, wang2025g3dlf}.
However, such approaches often fail to generalize to unseen scenarios that require broader real-world commonsense knowledge, and their decision-making processes typically lack transparency and interpretability.
To address these limitations, recent works have explored integrating LVLMs~\citep{yang2025qwen3, bai2025qwen25vl, OpenAI2024GPT4o, peng2023instruction} to assist navigation, either by leveraging stored knowledge or serving directly as decision backbones.
While fine-tuning LVLMs~\citep{lin2025evolvenav,zhou2024navgpt2, janusvln} on navigation datasets improves task adaptation, it sacrifices generalization and demands heavy resources.
Another growing line of research~\citep{zheng2024towards, wang2025dreamnav, mapgpt, liu2025msnav, han2025roomtour3d,lin2025navcot,zhang2024navid} has shifted toward training-free paradigms, where LVLMs act as embodied agents capable of performing navigation tasks without task-specific fine-tuning. 
Our work follows this direction, seeking to enhance the reasoning capability and interpretability of scene navigation systems through a training-free, dual-agent framework.

\subsection{Agentic Systems in Vision-Language Navigation}
The paradigm of LLMs has evolved from passive chatbots into autonomous agents capable of sophisticated reasoning, planning, and self-reflection~\citep{agent_survey, react, reflexion, tot}. While single agents show remarkable proficiency, they often face scalability and reasoning bottlenecks in long-horizon tasks\citep{mas_survey, Tran2025MultiAgentCM}. Consequently, recent research explores multi-agent collaboration, leveraging role-specialized agents and structured communication to partition cognitive workloads and enhance reliability~\citep{metagpt, mind, os_symphony}. These agentic systems prioritize autonomous decision-making, which is frequently augmented by memory structures or hierarchical control~\citep{kong2025mapagent}, to navigate complex environments.
In VLN, several methods adopt single-agent designs.
For instance, MapGPT~\citep{mapgpt} and MSNav~\citep{liu2025msnav} maintain topological maps to integrate long-term global context with local observations.
However, assigning both global reasoning and local action execution to a single agent often leads to entangled representations and suboptimal predictions.
To mitigate this, recent works have attempted to specialize roles through multi-agent systems. 
NavGPT~\citep{navgpt} combines a Navigation LLM with a Summarizer for trajectory compression and a Vision Foundation Model (VFM) for observation description, while CA-Nav~\citep{canav} utilizes an LLM-VFM hybrid to generate sub-instructions and value maps. Similarly, RAM~\citep{wei2025unseen} leverages LLMs and LVLMs for the semantic refinement of observations and instructions.
Moreover, DiscussNav~\citep{discussnav} employs five expert agents with distinct abilities, and DreamNav~\citep{wang2025dreamnav} extends this design to eight experts coordinated by four managers. 
While effective in principle, these designs require substantial computational resources (e.g., GPU usage or token costs) and introduce significant challenges in inter-agent coordination, with not all experts being strictly necessary.
In contrast, our work proposes a dual-agent framework that achieves efficient role separation, one agent for global planning and another for local execution—while maintaining lightweight and resource-efficient collaboration.

\section{Methodology}

In this section, we present DACo, a dual-agent collaborative system for Vision-and-Language navigation (VLN).
We first formalize the task and notations (Sec.~\S\ref{sec:formulation}).
Next, we introduce the overall framework (Sec.~\S\ref{sec:framework}) and describe its two core components: the \textbf{Global Agent} for intent-level planning (Sec.~\S\ref{sec:global_agent}) and the \textbf{Local Agent} for action-level execution (Sec.~\S\ref{sec:local_agent}).
Finally, we present the collaboration protocol that alternates between planning, execution, and re-planning, and provide the corresponding algorithmic workflow (Sec.~\S\ref{sec:agent_collaboration}).

\begin{figure*}[!th]
  \centering
  \includegraphics[width=\textwidth]{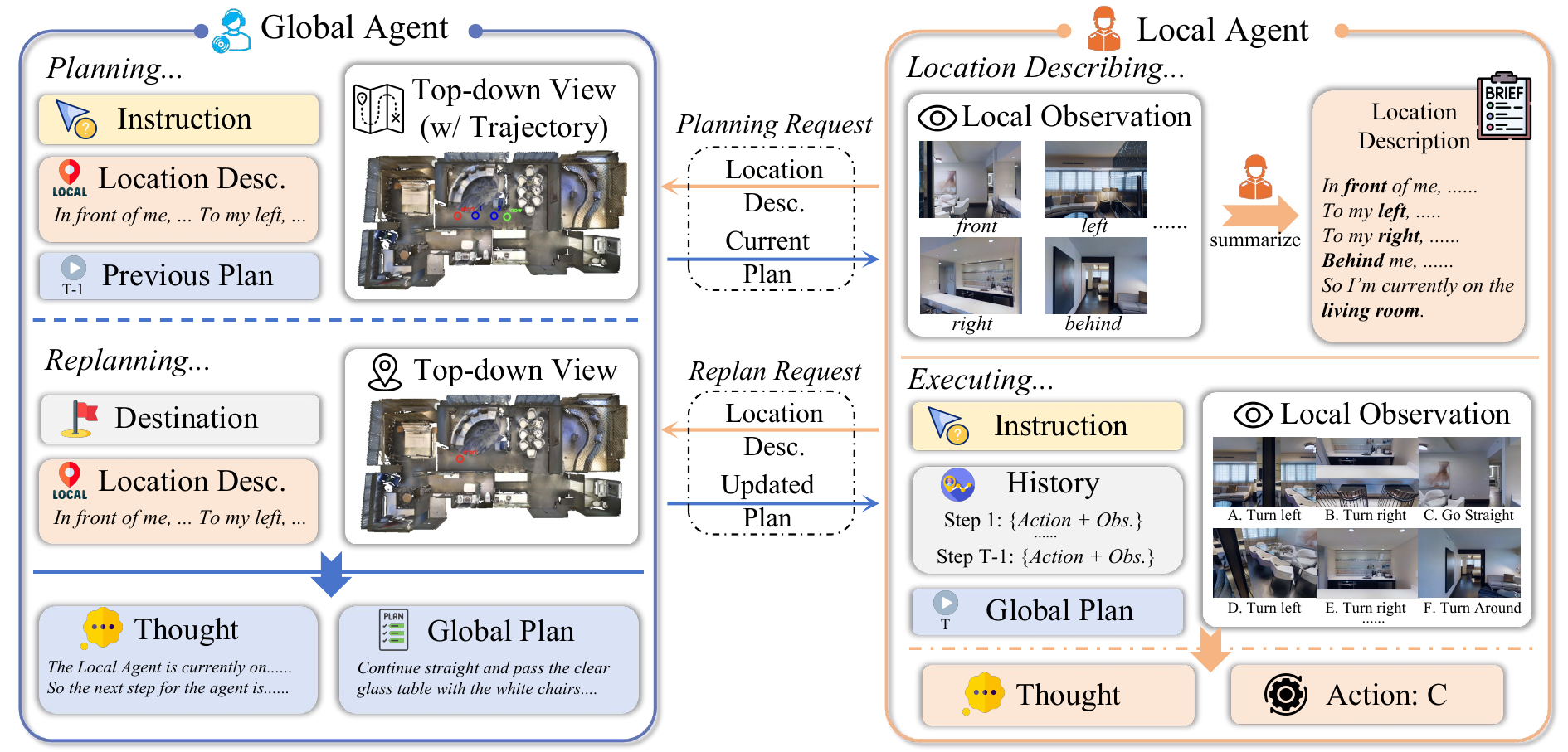}
  \vspace{-2mm}
    \caption{
    \textbf{Overview of the DACo framework.} Our system comprises two collaborative components:
    (1) \emph{Global Agent}: Acting as a high-level strategic planner, the Global Agent maintains a panoramic perspective by integrating current location descriptions, historical trajectories, and Top-down maps. It iteratively generates a dynamic global plan to guide the navigation. 
    (2) \emph{Local Agent}: Serving as the low-level executor, the Local Agent initiates each time step by synthesizing local observations into a concise environment description and issuing a planning or re-planning request. Upon receiving the global guidance, it grounds the high-level plan into primitive navigation actions to interact with the environment.
    }

  \label{fig:method_overview}
\end{figure*}

\subsection{Problem Formulation}
\label{sec:formulation}

VLN aims to enable an embodied agent to follow natural language instructions and reach a target location within a 3D environment. 
Formally, given a textual instruction $I$ and a 3D scene (e.g., Matterport3D~\citep{mattersim,Matterport3D}), the agent's objective is to generate a sequence of viewpoint-action pairs $\{(p_1, a_1), (p_2, a_2), ..., (p_T, a_T)\}$ which drives the agent, parameterized by a policy $\pi$, from the starting viewpoint $p_1$ to the goal location specified by $I$.
At each viewpoint $p_t$, the action $a_t$ is determined by:
\begin{equation}
   a_{t} = \pi(\mathcal{A}_t, O_{t}, I, \{(p_1, a_1), \cdots,(p_{t-1}, a_{t-1})\}, p_t)
\end{equation}
where $p_t$ denotes the current viewpoint and $a_{t}$ is the selected action that moves the agent to the next viewpoint $p_{t+1}$.
The local observation $O_t$ represents a panoramic RGB view captured at $p_t$: $O_t = \{o_{t, i}\}_{i=1}^{N}$, where each $o_{t, i}$ corresponds to an RGB image from a specific azimuth and elevation. 
This panoramic observation provides the agent with a comprehensive visual context for local surroundings at the current viewpoint.
The candidate action space $\mathcal{A}_t$ specifies all navigable directions from $p_t$ provided by the 3D simulator, and can be formulated as:
\begin{equation}
\mathcal{A}_t = \{\, \langle  a_{t,1}, p_{t+1,1}\rangle, \langle a_{t,2}, p_{t+1,2}\rangle, \dots, \langle a_{t,M}, p_{t+1,M} \rangle \,\} \cup \{\texttt{STOP}\},
\label{eq:action_space}
\end{equation}
where each pair $\langle a_{t,m}, p_{t+1,m}\rangle$ corresponds to an executable navigation action and its associated next viewpoint, $M$ is the size of the action space, and \texttt{STOP} indicates task termination.

\subsection{DACo}
\label{sec:framework}

\subsubsection{Overview}
The overview framework of DACo is illustrated in Fig.~\ref{fig:method_overview}, which consists of two agents operating within a closed-loop reasoning framework. 
At each navigation viewpoint $p_t$, DACo proceeds through the following closed workflow:
First, the \textbf{Local Agent} $\mathcal{F}_{\text{local}}$ generates a textual description from local observations and sends a \texttt{Planning} request to the \textbf{Global Agent} $\mathcal{F}_{\text{global}}$. 
The Global Agent then analyzes the current position $p_t$, trajectory history $H_t$, top-down scene view $\mathcal{B}$, and instruction $I$ to produce a current global plan $\Pi_t$.
Next, the Local Agent selects an executable action $a_t \in \mathcal{A}_t$ using its current panoramic observation $O_t$ and the received current plan $\Pi_t$. 
If inconsistency or uncertainty is detected (e.g., the plan is deemed unreachable or additional guidance is required), the Local Agent issues a \texttt{Replan} request to the Global Agent.
This design establishes a complementary division of labor: the Global Agent maintains long-term consistency with the instruction semantics and global scene structure, while the Local Agent adapts flexibly to dynamic and ambiguous local contexts. 
The overall system can be formulated as:
\begin{equation}
A = \pi\left(\mathcal{F}_{\text{global}}, \mathcal{F}_{\text{local}}, I, O, \mathcal{B} \right),
\label{eq:daco-overview}
\end{equation}
where $\mathcal{F}_{\text{global}}$ and $\mathcal{F}_{\text{local}}$ refer to the Global and Local Agents, respectively, and $\mathcal{B}$ provides global spatial context for high-level reasoning.

The following sections describe each component in detail and explain how the two agents collaborate through iterative planning, execution, and re-planning cycles.

\subsubsection{Global Agent}
\label{sec:global_agent}
The Global Agent functions as the high-level planner within DACo. 
Its primary role is to interpret navigation instructions and construct a global understanding of the environment, thereby enabling the generation of structured multi-step plans that guide the overall navigation process.
In this section, we introduce the key components of its prompt design and prediction mechanism.

\textbf{Global Observation.} 
To represent the global environment, we leverage the top-down scene view $\mathcal{B}$, as illustrated in Fig.~\ref{fig:method_overview}, which provides a compact and structured visual abstraction for high-level planning.
To facilitate the interpretation of the Local Agent’s past movements and enable reasoning about navigation progress, inspired by recent studies on \textbf{Visual Prompt}~\cite{vp_survey, os_symphony},
we incorporate the historical trajectory $H_t = \{(p_1, a_1), \cdots,(p_{t-1}, a_{t-1})\}$ on $\mathcal{B}$. 
Specifically, we overlay the coordinates of viewpoints in $H_t$ using distinct colors to indicate different stages: red marks the starting viewpoint, blue denotes intermediate viewpoints, and green represents the current viewpoint.
The coordinates of each viewpoint are obtained from preprocessed data.  
Finally, we obtain the marked top-down view $\tilde{\mathcal{B}}_t$. 

The resulting top-down scene representation enables the Global Agent to perceive the global spatial layout effectively. 
Although this representation is inherently coarse-grained, it provides sufficient contextual information for intent-level planning and long-term decision making.
Furthermore, our visual prompt design efficiently compresses historical trajectories, eliminating the need for redundant API calls and establishing a streamlined yet informative communication channel between the dual agents. 
This advantage in context management is particularly beneficial in long-horizon navigation tasks, where maintaining a compact yet comprehensive history is crucial for sustained performance.

\textbf{Plan Development.} 
Guided by the aforementioned global observations, the Global Agent formulates an executable trajectory from a high-level strategic perspective, which is subsequently transmitted to the Local Agent for fine-grained execution. 
Specifically, at each reasoning step $t$, the Global Agent receives four inputs: (1) the marked top-down scene image $\tilde{\mathcal{B}}_t$, 
(2) the navigation instruction $I$, 
(3) the global plan $\Pi_{t-1}$ generated at the previous step, and 
(4) the location description $\text{obs}_{\text{local}}$. 
Given the multimodal context, the global agent outputs a structured and dynamically updated global plan $\Pi_t = \{g_1, g_2, ..., g_K\}$, 
where each subgoal $g_i$ represents a semantically grounded intermediate landmark (e.g., ``pass the glass table'', ``turn left at the hallway''). 
Each subgoal implicitly encodes the intended spatial progression of the agent’s future movement.
Formally, the Global Agent is defined as:
\begin{equation}
\Pi_t = \mathcal{F}_{\text{global}}(I, \tilde{\mathcal{B}}_t, \Pi_{t-1}, \text{obs}_{\text{local}}),
\label{eq:global-plan}
\end{equation}
where $\text{obs}_{\text{local}}$ denotes the textual description of the local observation.

\subsubsection{Local Agent} 
\label{sec:local_agent}

While the Global Agent determines \emph{what to achieve}, the Local Agent decides \emph{how to act} within the current visual context.  
It is responsible for synthesizing the global plan, the original instruction, and the navigation history to interact with the environment simulator.  
At each time step, the Local Agent selects an optimal action from the candidate set provided by the simulator and advances to the corresponding viewpoint.  
In this section, we delineate the core components of the local prompt and detail the action execution process.

\textbf{Local Observation.}
At each viewpoint $p_t$, the Local Agent receives a set of local observations $O_t = \{o_{t,i}\}_{i=1}^{36}$, where $o_{t,i}$ denotes the visual context of the $i$-th viewpoint. 
Specifically, these 36 images are obtained by sampling 12 azimuth angles at each of three elevation levels ($-30^\circ$, $0^\circ$, and $30^\circ$), ensuring comprehensive environmental coverage.  
In addition, the agent is provided with the candidate action space $\mathcal{A}_t$ of size $M$, as defined in Eq.~\ref{eq:action_space}, together with the global plan
$\Pi_t = \{g_1, g_2, \ldots, g_K\}$ generated by the Global Agent.

Although the Global Agent generates a step-by-step executable path, we additionally incorporate the original instructions into the Local Agent’s input for two primary reasons. 
First, the synergy between the raw instructions and the global plan facilitates a more nuanced alignment with the user's intent. 
Second, while we have implemented a re-planning mechanism to resolve discrepancies between the global plan and local observations, we impose a re-planning quota to maintain computational efficiency and manage API costs.
When this budget is exhausted and the Global Agent’s guidance remains unreliable, the system adaptively falls back to a single-agent mode.  
In this mode, the Local Agent relies solely on local observations and the original instruction for navigation.  
This fallback mechanism serves as a critical safeguard, ensuring a stable performance floor and preventing cascading failures caused by accumulated LVLM errors.  
As a result, this multi-tier strategy substantially improves the overall robustness and stability of the framework.

Following previous works~\cite{mapgpt, liu2025msnav}, we record the navigation history $H_t = \{(p_1, a_1), \cdots,(p_{t-1}, a_{t-1})\}$ to facilitate progress monitoring and execution awareness.  
These actions are formatted as:
\begin{center}
    $step (1): \{a_1\}, \dots, step (t-1): \{a_{t-1}\}$.
\end{center}

\textbf{Action Execution.}
Given the local observations described above, the local reasoning process is formally defined as:
\begin{equation}
\langle a_t, p_{t+1} \rangle = \mathcal{F}_{\texttt{local}}(O_t, I, \Pi_t, H_t, \mathcal{A}_t), \quad \langle a_{t,*}, p_{t+1,*}\rangle \in \mathcal{A}_t.
\label{eq:local-decision}
\end{equation}
where the Local Agent selects an executable action $a_t$ and its corresponding next viewpoint $p_{t+1}$ from the candidate set $\mathcal{A}_t$.
Once the action $a_t$ is generated, it is executed in the simulator, which updates the agent’s state to the next viewpoint $p_{t+1}$ and initiates a new navigation cycle.  
As the primary interface for interacting with the environment, the Local Agent’s reasoning process is governed by a set of carefully designed protocols to ensure navigational consistency and execution reliability.  
Detailed prompt designs and interaction rules are provided in Sec.~\S\ref{prompt_details}.

\subsubsection{Agent Collaboration}
\label{sec:agent_collaboration}
\begin{figure*}[htbp]
  \centering
  \includegraphics[width=\textwidth]{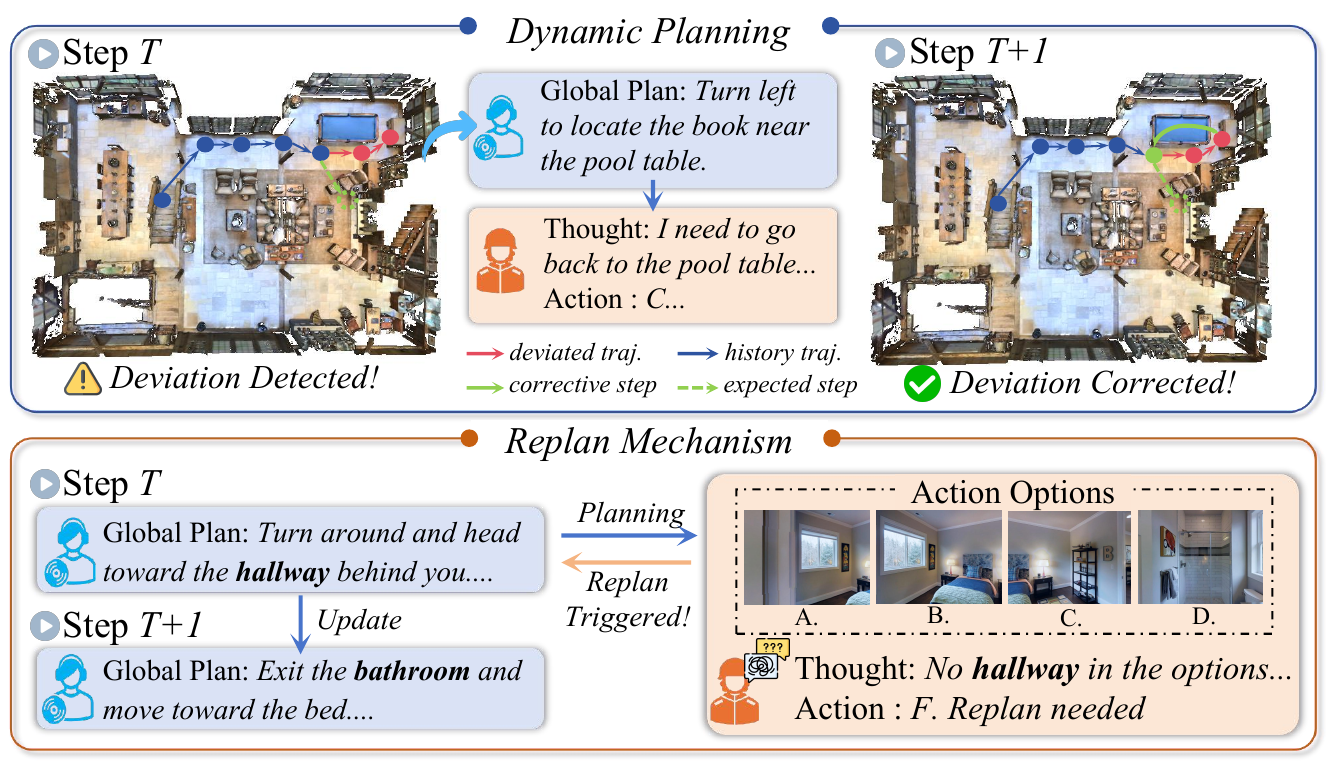}
  \vspace{-2mm}
    \caption{\textbf{The illustration of the self-correction capability inherent in DACo system.} Our framework achieves robust self-correction through two complementary processes: (1) \emph{Dynamic Planning}: The Global Agent monitors the Local Agent's trajectory via a top-down view; upon detecting a deviation, it adaptively refines the global plan to rectify the path. (2) \emph{Replan Mechanism}: The Local Agent cross-checks the global plan's validity against its immediate visual observations. If an inconsistency is detected (e.g., a missing landmark), a re-planning cycle is proactively triggered to resolve the discrepancy.
    }

  \label{fig:self_correcting}
\end{figure*}

The two agents collaborate through an iterative reasoning loop that alternates between top-down planning and bottom-up correction. 
Specifically, the Global Agent generates a \emph{dynamic plan} $\Pi_t$ at each time step, which is adaptively refined based on the Local Agent's latest trajectory. 
During execution, the Local Agent monitors the alignment between its observations and $\Pi_t$; any detected inconsistency triggers a \texttt{REPLAN} signal to initiate a new reasoning loop. These two mechanisms are illustrated in Fig.~\ref{fig:self_correcting}.

\textbf{Dynamic Planning Request: Local $\rightarrow$ Global.} 
At each viewpoint, the \textbf{Local Agent} first sends a planning request to the Global Agent along with a description of its current observation. 
Specifically, twelve images $\hat{O}_t = \{ o_{t,i}, \text{cap}_{t,i}\}_{i=1}^{12}$ are selected at an elevation angle of $0^\circ$, where each image is obtained from the simulator. 
Using the orientation angle provided by the simulator, we generate a short orientation caption ($\text{cap}_{t}$) for each image, such as ``Image 1 (in front of you)'', ``Image 2 (on your right)''.
We then pair each image with its corresponding caption to form the set $\hat{O}_t = \{\langle o_{t,i}, \text{cap}_{t,i}\rangle\}_{i=1}^{12}$.
Based on this set of images and captions, the Local Agent in this step can be regarded as a location descriptor, defined as:
\begin{equation}
\mathcal{F}_{\text{local}} \xrightarrow{\text{obs}_{\text{local}}=\mathcal{F}_{\text{local}}(\hat{O}_t)} \mathcal{F}_{\text{global}}
\label{eq:get-cur-loc}
\end{equation}

\textbf{Dynamic Planning Response: Global $\rightarrow$ Local.}
Once the Global Agent receives a \texttt{Planning} request, it generates a dynamic plan. Specifically, at each time step, the Global Agent is tasked with reformulating the global plan $\Pi_t$ according to the formal definition in Eq.~\ref{eq:global-plan}. The dynamic correction process is illustrated in the upper portion of Fig.~\ref{fig:self_correcting}, and defined as:
\begin{equation}
\mathcal{F}_{\text{global}} \xrightarrow{\Pi_t} \mathcal{F}_{\text{local}}
\end{equation}

The reason we increase the interaction frequency between the two agents is to improve the reliability of the planning process, ensuring that the Local Agent always operates with an up-to-date global context. 
This design serves to ensure that any trajectory drifts of the Local Agent are rectified in a timely manner.
By taking the global representation $\tilde{\mathcal{B}}_t$ as input at each time step, the Global Agent is capable of monitoring the Local Agent's trajectory from a global perspective. Consequently, if the Local Agent deviates from the intended path, the Global Agent can promptly adjust the high-level plan to rectify the trajectory and steer it back on track.

\textbf{Replan Request: Local $\rightarrow$ Global.}
In contrast to the standard planning flow, the replanning process reverses the direction of information exchange. 

Beyond the top-down guidance, we design a reciprocal verification protocol enabling the Local Agent to scrutinize the feasibility of the Global Plan based on its immediate environmental observations.
Due to the typically coarse-grained nature of global representations, discrepancies may arise between the Global Plan and the Local Agent's fine-grained observations. To address this, we implement a verification step during the execution phase, as shown in the lower portion of Fig.~\ref{fig:self_correcting}. At each time step, the Local Agent validates whether the Global Plan aligns with its current action candidates. If a landmark specified in the Global Plan is absent from the local action options, the Local Agent identifies a planning error and proactively triggers a \texttt{Replan} request. This initiation of a new reasoning loop ensures that the navigation remains grounded in the immediate environmental context.

Upon receiving this request, the Global Agent generates a new global plan $\Pi'_t$ from scratch, without relying on the previous plan $\Pi_{t-1}$. 
Since the original instruction is given relative to the starting point, while the Local Agent may have moved far away, the Global Agent first extracts the target location from the instruction, then marks the Local Agent’s current position on the top-down scene view as the new starting point. 
Using this updated context, along with the attached observation, a new plan is computed. 
The reasoning process is defined as:
\begin{equation}
\Pi'_t = \mathcal{F}_{\text{global}}(\tilde{\mathcal{B}}_t, \text{target}, \text{obs}_{\text{local}}),
\label{eq:re-plan}
\end{equation}

\textbf{Replan Response: Global $\rightarrow$ Local.}
Upon receiving a \texttt{REPLAN} request, the Global Agent generates a new global plan $\Pi'_t$ and transmits it to the Local Agent, which resumes navigation based on the updated plan.

The proposed collaboration mechanism is designed to balance global semantic consistency with local adaptability, enabling robust long-horizon navigation under zero-shot settings.  
Overall, our framework establishes a collaborative yet interdependent interaction between the two agents.  
Rather than operating in isolation, the agents continuously supervise and refine each other through iterative information exchange and coordinated decision making.
This architecture naturally induces an intrinsic self-correction mechanism, which effectively mitigates error accumulation and substantially enhances system reliability and robustness across complex multi-room and multi-level trajectories.  
The complete collaboration workflow is summarized in Algorithm.~\ref{alg:agent_collaboration}.

\begin{algorithm}[!t]
\caption{Agent Collaboration Workflow in DACo}
\label{alg:agent_collaboration}
\begin{algorithmic}[1]  
\Require{Instruction $I$, Initial viewpoint $p_0$}, Max action length $L$, A simulator $\texttt{sim}$
\Ensure{Action sequence $A = \{(p_1, a_1), (p_2, a_2), ..., (p_T, a_T)\}$}

\State Initialize Local Agent $\mathcal{F}_{\texttt{local}}$ and Global Agent $\mathcal{F}_{\texttt{global}}$
\State $t \gets 0$
\While{t $<$ L}
    \State $O_t \gets $ get visual observation from $\texttt{sim}$
    \State \textbf{(Local)} $\text{obs}_{\text{local}} \gets$ Generate local observation using Eq.~\ref{eq:get-cur-loc} and send the \texttt{PLANNING} request
    \State \textbf{(Global)} $\Pi_t \gets$ Generate or update global plan using Eq.~\ref{eq:global-plan} and send the \texttt{PLANNING} response
    \State \textbf{(Local)} $a_t \gets$ Generate action using Eq.~\ref{eq:local-decision}
    \If{$a_t$ = \texttt{REPLAN}}
        \State \textbf{(Local)} $\text{obs}_{\text{local}} \gets$ Generate local observation using Eq.~\ref{eq:get-cur-loc} and send the \texttt{REPLAN} request
        \State \textbf{(Global)} $\Pi'_t \gets$ Re-generate a new plan using Eq.~\ref{eq:re-plan} and send the \texttt{REPLAN} response
    \Else 
        \If{$a_t$ != \texttt{STOP}}
            \State Update position $p_t \gets p_{t+1}$
            \State $A \gets A\ \cup a_t$
            \State $t \gets t + 1$
        \Else
            \State break
        \EndIf
    \EndIf
\EndWhile \\
\Return Action sequence $A$
\end{algorithmic}
\end{algorithm}

\section{Experiments}
In this section, we carried out extensive experiments on several widely used Vision-Language Navigation benchmarks to answer the following research questions:
\begin{itemize}
    \item \textbf{RQ1}: \textbf{Comparative Performance.} 
    How does DACo perform relative to existing zero-shot state-of-the-art (SOTA) agentic frameworks across VLN tasks of varying difficulty and environmental complexity?

    \item \textbf{RQ2}: \textbf{Ablation Study}. 
    To what extent do the individual components and mechanisms of DACo contribute to its overall navigational success?

    \item \textbf{RQ3}: \textbf{Generalization Across Backbones.} 
    Does DACo consistently outperform baseline frameworks regardless of the underlying agentic backbone (e.g., GPT vs. Qwen series)?

    \textbf{RQ4}: \textbf{Operational Efficiency.}
    What are the computational overheads and API costs associated with DACo compared to representative baselines?
\end{itemize}

\subsection{Experimental Setup}

\textbf{Datasets.}
In this work, we evaluate the proposed method across three widely-adopted benchmarks: \textbf{R2R}~\citep{R2R}, \textbf{REVERIE}~\citep{reverie}, and \textbf{R4R}~\citep{R4R}. 
All three benchmarks are built upon the large-scale 3D indoor environment dataset Matterport3D~\citep{Matterport3D}, yet each emphasizes distinct challenges to represent varying levels of task complexity.
R2R serves as a foundational benchmark, providing fine-grained, step-by-step instructions where each task typically requires only 5 to 7 steps to complete.
In contrast, REVERIE utilizes the same ground-truth trajectories as R2R but replaces detailed guidance with concise, high-level instructions—often a single sentence describing a target object. 
This setup places a higher premium on the agent's spatial reasoning and goal inference capabilities. 
Finally, R4R is designed to assess long-horizon navigation. 
By concatenating multiple R2R paths into extended trajectories that often exceed ten steps, it necessitates sustained long-range planning. 
This diverse selection of benchmarks ensures a comprehensive evaluation of the agent from multiple perspectives, demonstrating the robustness and versatility of our framework.
Note that the original REVERIE task involves two components: navigating to the target object and grounding the referred object. 
Following prior work~\citep{mapgpt}, we focus exclusively on the navigation component and omit the grounding subtask.

\textbf{Baselines.} 
We compare our method against representative zero-shot scene navigation approaches, including NavGPT~\citep{navgpt}, DiscussNav~\citep{discussnav}, MapGPT~\citep{mapgpt}, and MSNav~\citep{liu2025msnav}. 
Among these methods, MSNav is the only baseline that is not fully training-free, as it fine-tunes the Qwen-3 model on REVERIE to enhance spatial reasoning.
Since it is closed-source, it is not feasible for us to reproduce its results on alternative model backbones.
To highlight the distinction, we mark MSNav in light-colored text in our result tables, as our primary focus is on fully training-free zero-shot methods.
In addition to zero-shot comparisons, we also include state-of-the-art methods under alternative settings, such as those that rely on task-specific training (\texttt{Train}), pre-trained vision–language models (e.g., ViT, BERT) (\texttt{Pretrain}), and fine-tuned large language models (\texttt{LLM-Ft}).
Notably, despite operating in a fully training-free regime, our method still surpasses several approaches that require training, highlighting the effectiveness and generality of our design.

\textbf{Implementation.}
We use GPT-4o~\citep{OpenAI2024GPT4o} for both the Global and Local agents, operating in a purely zero-shot setting without fine-tuning. 
For experiments with open-source models, we deployed Qwen2.5-VL-32B-Instruct~\citep{bai2025qwen25vl} and Qwen3-VL-8B-Instruct~\citep{bai2025qwen3vltechnicalreport} on two A40 GPUs, each equipped with 40 GB of VRAM. 
The top-down scene images and the viewpoint coordinate data are derived from the publicly available WayDataset~\citep{whereareyou}. 

Moreover, in experiments, we set the maximum step limit to 15 for the R2R and REVERIE benchmarks, and extended it to 30 for the R4R long-horizon tasks. This configuration was consistently applied to both our proposed method and all re-implemented baselines. 
For the LLM/LVLM API configurations, we set the sampling temperature to 0 to minimize output stochasticity and capped the maximum generation length at 1000 tokens. Meanwhile, to balance testing efficiency and API cost, we do not allow the local agent to request re-planning indefinitely. We limit the local agent to at most one replan request.

\textbf{Evaluation Metrics.}
We adopt standard evaluation protocols for vision-and-language navigation and report four widely used metrics:
(1) \textbf{Navigation Error (NE)} — the average shortest-path distance (in meters) between the agent's final position and the target location; 
(2) \textbf{Success Rate (SR)} — the percentage of navigation episodes in which the agent terminates within 3m of the goal; 
(3) \textbf{Oracle Success Rate (OSR)} — the percentage of episodes in which the closest point along the agent’s trajectory falls within 3m of the goal, assuming an oracle stopping policy; and 
(4) \textbf{Success weighted by Path Length (SPL)} — a composite measure balancing success and trajectory efficiency, computed as $\frac{1}{N} \sum_i S_i \frac{L_i}{\text{max}(L_i ,P_i)}$, where $S_i$ is the success indicator for episode $i$, $L_i$ the shortest-path distance to the goal, and $P_i$ the actual trajectory length.

\subsection{Main Results (RQ1)}

\begin{table*}[!t]
\centering
\caption{Results on 72 various scenes of the R2R dataset. MapGPT is reproduced using GPT-4o for fair comparison. Bold denotes the best result in the training-free method. Light color denotes the method with the fine-tuning stage}
\label{tab:R2R-72-results}
\fontsize{9}{10}\selectfont
\setlength{\tabcolsep}{3.8mm}
\begin{tabular}{l l | c c c c }
\toprule
\textbf{Methods} & \textbf{Backbone} & \textbf{NE$\downarrow$} & \textbf{OSR$\uparrow$} & \textbf{SR$\uparrow$} & \textbf{SPL$\uparrow$} \\
\midrule
NavGPT~\citep{navgpt} & GPT-3.5 & 8.02 & 26.4 & 16.7 & 13.0 \\
DiscussNav~\citep{discussnav} & GPT-4 & 6.30 & 51.0 & 37.5 & 33.3 \\
MapGPT~\citep{mapgpt} & GPT-4o & \textbf{5.45} & 55.1 & 44.6 & 35.4 \\
\textcolor{gray}{MSNav~\citep{liu2025msnav}} & \textcolor{gray}{Qwen-SP} & \textcolor{gray}{5.02} & \textcolor{gray}{63.9} & \textcolor{gray}{50.9} & \textcolor{gray}{42.6} \\
DACo~(Ours) & GPT-4o & 5.86 & \textbf{63.7} & \textbf{50.5} & \textbf{39.7} \\
\bottomrule
\end{tabular}
\end{table*}

\subsubsection{Results on the Room-to-Room Dataset}

\textbf{Various Scenes.} We follow the standard subset as in MapGPT's experiment (72 scenes, 216 samples) to evaluate the zero-shot performance across various scenes. For the sake of fair comparison while reducing API-related costs, we re-evaluated the previous open source SOTA method, MapGPT, using GPT-4o. Results for other approaches are sourced directly from their respective original papers~\citep{navgpt, discussnav, lin2025navcot, liu2025msnav}. As shown in Table~\ref{tab:R2R-72-results}, DACo surpasses the previous methods in SR, OSR, and SPL metrics, achieving an SR of 50.5\% and an OSR of 63.7\%. Meanwhile, compared to the MapGPT, our method exhibits a slight degradation in the NE. We attribute this to the design of our dual-agent workflow, and we refer to this phenomenon as "error accumulation". 
In our framework, each navigation step of the local agent depends not only on the instruction and egocentric visual observations, but also on the planning provided by the global agent. Consequently, in some cases, an erroneous plan from the global planner may propagate to the local agent, leading to deviated trajectories and thus resulting in increased NE. Nevertheless, we argue that the most critical metrics for navigation are SR and OSR. The significant improvements in these two metrics demonstrate that our approach effectively enhances task completion in the majority of scenarios. In addition, our method performs almost on par with MSNav~\cite{liu2025msnav}, a method using a fine-tuned agent, showing only a negligible decline, which further demonstrates the effectiveness of our approach.

\begin{table*}[!t]
\centering
\caption{Results on R2R validation unseen set. Bold denotes the best result in the training-free method. * denotes the zero-shot version provided by the original paper. Light color denotes the method with the fine-tuning stage. }
\label{tab:R2R-val-unseen-results}
\fontsize{9}{10}\selectfont
\setlength{\tabcolsep}{4.0mm}
\begin{tabular}{l l l l l l}
\toprule
\textbf{Settings} & \textbf{Methods} & \textbf{NE}$\downarrow$ & \textbf{OSR}$\uparrow$ & \textbf{SR}$\uparrow$ & \textbf{SPL}$\uparrow$ \\
\midrule
\multirow{3}{*}{Train} & Seq2Seq~\citep{R2R} & 7.81 & 28 & 21 & - \\
      & Speaker~\citep{fried2018speaker}  & 6.62 & 45 & 35 & - \\
      & EnvDrop~\citep{tan2019learning} & 5.22 & - & 52 & 48 \\
\midrule
\multirow{8}{*}{Pretrain} & PREVALENT~\citep{hao2020prevalent} & 4.71 & - & 58 & 53 \\
         & RecBERT~\citep{hong2021vln} & 3.93 & 69 & 63 & 57 \\
         & HAMT~\citep{chen2021hamt} & 2.29 & 73 & 66 & 61 \\
         & DUET~\citep{chen2022think} & 3.31 & 81 & 72 & 60 \\
         & LangNav~\citep{pan2023langnav} & - & - & 43 & - \\
         & ScaleVLN~\citep{wang2023scaling}  & 2.09 & 88 & 81 & 70 \\
         & GOAT~\citep{wang2024vision} & 2.40 & 85 & 78 & 68\\
         & GOAT-ATENA~\citep{atena} & 2.27 & - & 79 & 69 \\
\midrule
\multirow{4}{*}{LLM-Ft} & NavCoT~\citep{lin2025navcot} & 6.26 & 48 & 40 & 37 \\
         & NaviLLM-~\citep{zheng2024towards} & 3.51 & - & 67 & 59 \\
         & NavGPT-2~\citep{zhou2024navgpt2} & 3.37 & 74 & 68 & 56 \\   
         & EvolveNav~\citep{lin2025evolvenav} & 3.15 & - & 71 & 63 \\
\midrule
\multirow{6}{*}{Zero-shot} 
   & NavGPT~\citep{navgpt} & 6.46 & 42 & 34 & 29 \\
   & DiscussNav~\citep{discussnav} & \textbf{5.32} & 61 & 43 & \textbf{40} \\
   & MapGPT~\citep{mapgpt} & 5.63 & 58 & 44 &  35 \\
   & NavCoT*~\citep{lin2025navcot} & 6.95 & 45 & 32 & 29 \\
   & \textcolor{gray}{MSNav~\citep{liu2025msnav}} & \textcolor{gray}{5.24} & \textcolor{gray}{65} & \textcolor{gray}{46} &  \textcolor{gray}{40} \\
   &  DACo~(Ours) & 5.94 & \textbf{67} & \textbf{48} & 39 \\
\bottomrule
\end{tabular}
\end{table*}

\textbf{Validation Unseen Set.} To comprehensively evaluate our work, we further compare the performance on a larger dataset, i.e., the full validation unseen set of R2R. This set has 11 different scenes, 783 trajectories, and 2,349 instructions (3 instructions per trajectory).
As shown in Table ~\ref{tab:R2R-val-unseen-results}, our method outperforms all zero-shot methods in SR and OSR, even including MSNav~\cite{liu2025msnav}. 
Similarly, our method has a slightly higher NE than the previous SOTA, and the reason has already been explained earlier. Furthermore, we observe that our method has suboptimal SPL. After our analysis, this is due to our dynamic planning and replanning mechanism. We incorporate these two mechanisms to allow the local agent to receive updated global instructions at different execution steps. 
While this mechanism enables the agent to recover from erroneous trajectories and eventually reach the target in challenging scenarios, it may result in slightly longer execution paths compared to the optimal route, thereby lowering the SPL metric. 
We argue that this modest reduction in path efficiency is justified by the substantial improvement in task success, which more directly reflects the primary objective of VLN. 
Notably, under the training-free zero-shot setting, our method also surpasses several trained approaches, such as NavCoT.

\begin{table*}[!t]
\centering
\caption{Results on REVERIE validation unseen set. Bold denotes the best result in the training-free method. Methods in ``Train", ``Pretrain", and ``LLM-Ft" settings are evaluated on the full set. NavGPT and MapGPT are reproduced using GPT-4o.}
\fontsize{9}{10}\selectfont
\setlength{\tabcolsep}{3.5mm}
\begin{tabular}{l l l l l l}
\toprule
\textbf{Settings} & \textbf{Methods} & \textbf{NE$\downarrow$} & \textbf{OSR$\uparrow$} & \textbf{SR$\uparrow$} & \textbf{SPL$\uparrow$} \\
\midrule
\multirow{3}{*}{Train} 
      & Seq2Seq~\citep{R2R} & - & 8.1 & 4.2 &  2.8 \\
      & RCM~\citep{wang2019reinforced}  & - & 14.2 &  9.3 & 7.0 \\
      & FAST-MATTN~\citep{reverie} & - & 28.2 & 14.4 & 7.2 \\
\midrule
\multirow{8}{*}{Pretrain}
         & RecBERT~\citep{hong2021vln} & - & 35.0 & 30.7 & 24.9 \\
         & HAMT~\citep{chen2021hamt} & - & 35.4 & 31.6 & 29.6 \\
         & DUET~\citep{chen2022think} & - & 51.0 & 46.9 & 33.7 \\
         & LAD~\citep{li2023lad} & - & - & 57.0 & 37.9 \\
         & ScaleVLN~\citep{wang2023scaling}  & - & 63.9 & 57.0 &  41.8 \\
         & GOAT~\citep{wang2024vision} & - & - & 53.4 & 36.7\\
         & GOAT-FSTTA~\citep{fstta} & - & 57.9 & 53.8 & 37.5 \\
         & GOAT-ATENA~\citep{atena} & - & 70.3 & 67.7 & 53.2 \\
\midrule
\multirow{2}{*}{LLM-Ft} & NavCoT~\citep{lin2025navcot} & - & 14.2 & 9.2 & 7.2 \\
         & NaviLLM~\citep{zheng2024towards} & - & - & - & 35.7 \\
\midrule
\multirow{3}{*}{Zero-shot} & NavGPT~\citep{navgpt} & 8.9 & 23.5 & 19.0 & 16.6 \\
   & MapGPT~\citep{mapgpt} & 8.8 & 44.0 & 30.0 & 21.2 \\
   & DACo~(Ours) & \textbf{7.8} & \textbf{56.0} & \textbf{36.5} & \textbf{25.2} \\
\bottomrule
\end{tabular}
\label{tab:REVERIE-val-unseen-results}
\vspace{-4mm}
\end{table*}
\subsubsection{Results on the REVERIE Dataset}

\textbf{Validation Unseen Set.}
Previous work, due to evaluation efficiency and API cost considerations, evaluated on a subset of the REVERIE unseen validation set. 
However, the specific subsets used in prior studies have not been publicly released.
To ensure comparability under similar constraints, we randomly sample 200 instances from the validation set for evaluation.
As reported in Table~\ref{tab:REVERIE-val-unseen-results}, our method demonstrates strong performance on REVERIE, outperforming all methods under the zero-shot setting, as well as all methods in the “Train” setting, two methods in the “Pretrain” setting, and one method in the “LLM-Ft” setting. 
Compared with MapGPT, DACo achieves a 6.5\% absolute improvement in SR and a 12\% improvement in OSR, exceeding the gains observed on R2R.
These results highlight the effectiveness of explicitly introducing global planning. Unlike R2R, REVERIE provides only brief instructions without a predefined route, increasing task ambiguity. 
In such cases, the global commander generates a structured path plan based on BEV representations, compensating for the limited instruction specificity and providing the local operative with a coherent global context. 
This structured guidance enhances spatial reasoning and robustness in complex navigation scenarios.

\begin{table*}[!th]
\centering
\caption{Comparison with zero-shot state-of-the-art methods on R4R validation unseen subset. All methods are implemented with Qwen3-VL-8B for fair comparison.}
\fontsize{9}{10}\selectfont
\setlength{\tabcolsep}{4.5mm}
\begin{tabular}{l l l l l}
\toprule
\textbf{Methods} & \textbf{NE$\downarrow$} & \textbf{OSR$\uparrow$} & \textbf{SR$\uparrow$} & \textbf{SPL$\uparrow$} \\
\midrule

NavGPT~\citep{navgpt} & 9.90 & 44.0 & 15.0 & 14.1 \\
MapGPT~\citep{mapgpt} & 8.94 & 54.8 & 15.1 & 13.0 \\
DACo~(Ours) & \textbf{8.91} & \textbf{55.0} & \textbf{20.5} & \textbf{17.4} \\
\bottomrule
\end{tabular}
\label{tab:R4R-results}
\end{table*}

\subsubsection{Results on the R4R Dataset}

In this section, we evaluate the long-horizon navigation capability of DACo on the R4R benchmark.
We compare our approach with two representative zero-shot agentic frameworks, NavGPT and MapGPT, which exemplify the multi-agent and single-agent paradigms, respectively.
Due to computational and inference cost constraints, evaluation is conducted on a randomly sampled subset of 200 tasks.
As shown in Table~\ref{tab:R4R-results}, zero-shot frameworks generally exhibit substantial performance degradation on long-horizon tasks, with markedly lower success rates compared to their results on R2R and REVERIE.
This decline likely stems from the difficulty of maintaining coherent reasoning over extended context windows, a known limitation of current LLMs and LVLMs in long-range decision-making.
Against this backdrop, DACo consistently outperforms both MapGPT and NavGPT.
In particular, DACo achieves an absolute improvement of approximately 5\% in success rate (corresponding to a 33\% relative gain) over the strongest baseline, while also leading across other evaluation metrics.
These results suggest that explicitly incorporating global spatial planning enhances reasoning stability and effectiveness in long-horizon navigation scenarios.

\subsection{Ablation Study (RQ2)}

\begin{table}[!t]
\centering
\caption{Ablation on 72 various scenes of the R2R dataset. }
\fontsize{9}{10}\selectfont
\setlength{\tabcolsep}{6.5mm}
\begin{tabular}{cc|cccc}
\toprule
\textbf{Plan Style} & \textbf{Replan}  & \textbf{NE$\downarrow$} & \textbf{OSR$\uparrow$} & \textbf{SR$\uparrow$} & \textbf{SPL$\uparrow$} \\
\midrule
    $\times$ & $\times$ & 6.44 & 63.5 & 45.7 & 35.1 \\
    static & $\checkmark$ & 5.79 & 64.3 & 46.9 & 36.4 \\
    dynamic & $\times$ & 6.2 & 61.6 & 47.5 & 38.2 \\
    dynamic & $\checkmark$ & \textbf{5.86} & \textbf{63.7} & \textbf{50.5} & \textbf{39.7} \\
\bottomrule
\end{tabular}
\label{tab:ablation-results}
\end{table}

As shown in Table \ref{tab:ablation-results}, we explore the effectiveness of different collaboration mechanism designs between two agents on 72 various scenes of the R2R dataset. 

\textbf{The Effect of Different Global Plan Styles.}
We analyze the role of the global agent and examine how the interaction frequency between the two agents affects performance.
We consider two variants: \emph{Static Plan}, where the global agent generates a single plan prior to navigation, and \emph{Dynamic Plan}, where the global agent updates the plan at each step based on the current observations and trajectory history.
As shown in the experimental results, dynamic planning consistently outperforms static planning, and both variants surpass the configuration without a global agent.
This comparison highlights the importance of integrating global guidance, as well as the benefit of iterative plan refinement during execution.
A qualitative example is provided in the upper panel of Fig.~\ref{fig:self_correcting}.
With dynamic planning, deviations made by the Local Agent are corrected promptly through updated global guidance.
In contrast, without such iterative adjustment, the agent continues along a misaligned trajectory (highlighted in red), ultimately resulting in navigation failure.

\textbf{The Effect of Replan Mechanism.} We found that introducing the Replan mechanism improves SR by 3.0\% and OSR by approximately 2.1\%, demonstrating the effectiveness of the Replan mechanism. Moreover, this result indicates that the agent possesses the ability to judge when it should seek assistance. Furthermore, we tested the replan ratio on both R2R and REVERIE, and the results show that the model's replan ratio on REVERIE is \textbf{twice} that on R2R. This suggests that agents are more inclined to seek help and collaborate on more complex tasks.

\subsection{The Impact of Agentic Backbone (RQ3)}

\begin{figure}[!t]
    \centering
    \includegraphics[width=0.55\textwidth]{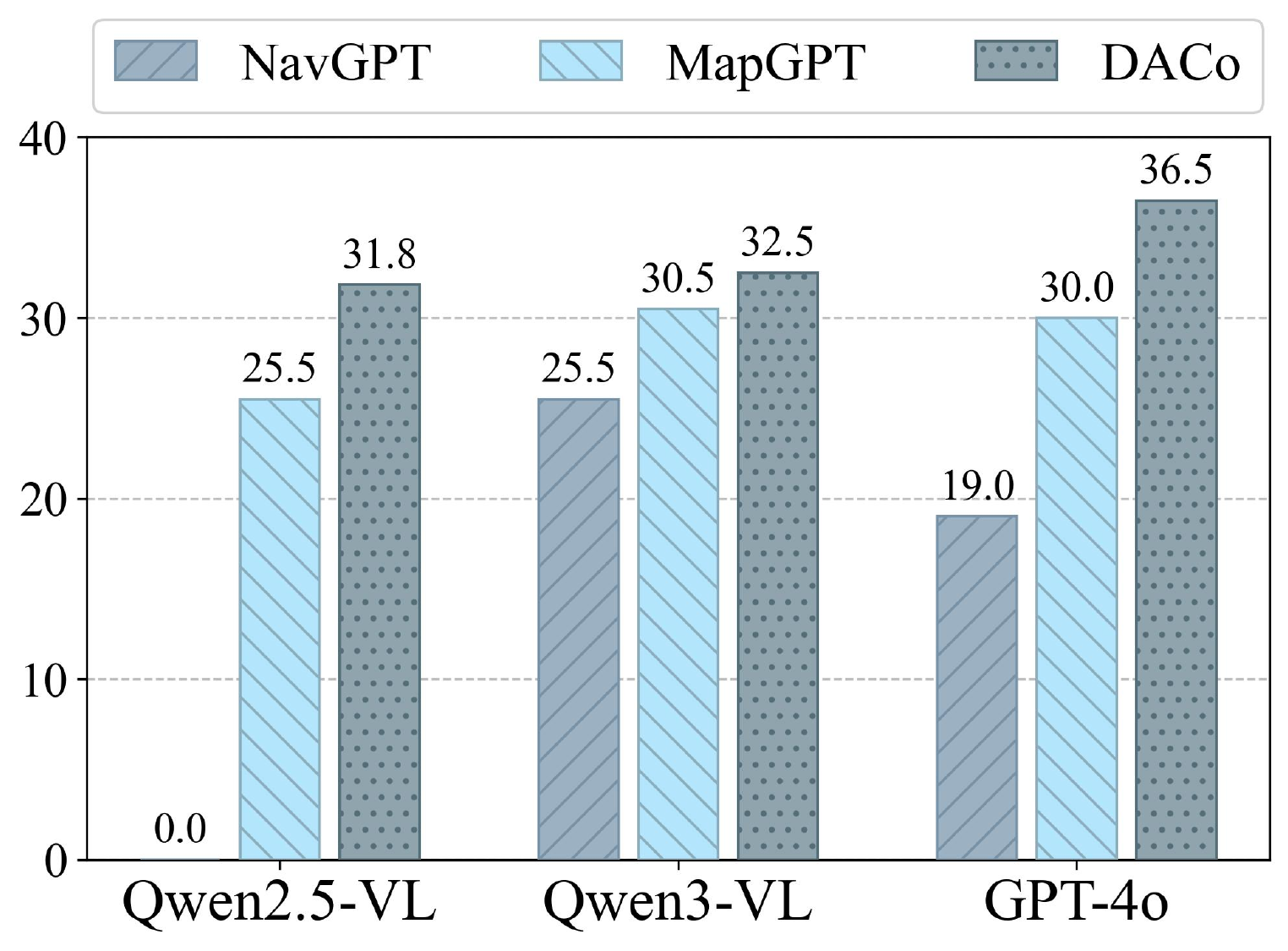}
    \vspace{-2mm}
   \caption{Impact of Agent Backbone. All methods are implemented using Qwen2.5-VL-32B, Qwen3-VL-8B, and GPT-4o. The zero success rate of NavGPT on Qwen2.5-VL is primarily due to catastrophic format parsing failures.}
    \label{fig:backbone-result}
\end{figure}

To investigate the impact of different agentic backbones on navigation performance, we further evaluated DACo, MapGPT, and NavGPT on the REVERIE test set using Qwen3-VL-8B~\cite{bai2025qwen3vltechnicalreport} and Qwen2.5-VL-32B~\cite{bai2025qwen25vl}.
The results in Fig.~\ref{fig:backbone-result} reveal a clear correlation between backbone capability and navigation performance, with stronger models generally achieving higher success rates.
Notably, when using Qwen2.5-VL-32B, NavGPT fails to produce valid outputs, resulting in a zero success rate. 
This behavior can be attributed to its strict output formatting constraints, which reduce robustness when the backbone does not precisely follow predefined templates.
In contrast, DACo maintains stable and competitive performance across different backbones, indicating that its architectural design does not rely on model-specific prompting behaviors and generalizes well across heterogeneous LVLMs.
Furthermore, DACo equipped with Qwen backbones surpasses MapGPT powered by GPT-4o.
This demonstrates that our framework can achieve superior performance using fully open-source models, highlighting both its architectural effectiveness and its practical advantages in terms of accessibility and deployment cost.

\begin{table}[!t]
\centering
\caption{Analysis of efficiency and API cost. All experiments are carried out with Qwen3-VL-8B on the REVERIE subset. \textit{Time} and \textit{Prompt/Completion Token} denote the average per task, while \textit{Latency} refers to the average per call.}
\fontsize{9}{10}\selectfont
\setlength{\tabcolsep}{3.5mm}
\begin{tabular}{l|llll}
\toprule
\textbf{Method} & \textbf{Time$\downarrow$} & \textbf{Prompt Token$\downarrow$} & \textbf{Completion Token$\downarrow$} & \textbf{Latency$\downarrow$} \\
\midrule
NavGPT~\citep{navgpt} & 106.14 & $\mathbf{3.56 \times 10^4}$ & $2.79 \times 10^3$ & 7.0 \\
MapGPT~\citep{mapgpt} & \textbf{63.24} & $5.33 \times 10^4$ & $\mathbf{1.76 \times 10^3}$ & \textbf{6.1} \\
DACo w/ static plan & 95.48 & $4.72 \times 10^4$ & $1.99 \times 10^3$ & 6.8 \\
\bottomrule
\end{tabular}
\label{tab:api-cost-results}
\vspace{-4mm}
\end{table}
\subsection{Trade-off Between Accuracy and Cost. (RQ4)}
While previous sections demonstrate the performance advantages of DACo, we further analyze its computational overhead and API consumption to assess practical efficiency.

In our ablation study, we propose two variants of global planning: dynamic and static. While dynamic planning demonstrates better performance, it inevitably incurs higher token consumption due to the requirement of agent interaction at every action step. In the case of static planning, due to the reduced number of agent interactions, the number of API calls is nearly half that of dynamic planning, leading to substantial savings in both computational cost and processing time, while incurring only a marginal performance degradation of approximately 3\% (as shown in Table \ref{tab:ablation-results}). Consequently, in this subsection, DACo with a static plan is introduced as a balanced variant, optimized for both efficiency and performance, to serve as a comparative baseline.

Table~\ref{tab:api-cost-results} reports execution time, token usage, and API cost for a lightweight version of DACo compared with NavGPT and MapGPT.
In terms of runtime, the single-agent design (MapGPT) naturally benefits from minimal coordination overhead.
Nevertheless, DACo significantly reduces latency relative to the multi-agent framework (NavGPT), with an average improvement of approximately 10 seconds.
In terms of token consumption, although NavGPT records the lowest prompt token count, this is primarily attributed to its reliance on pre-processed visual information, which bypasses the substantial overhead of online visual token inputs. Given NavGPT's significant performance gap compared to MapGPT and DACo, we argue that this token-saving strategy comes at an unacceptable cost to accuracy. Notably, DACo achieves a lower prompt token count than the single-agent baseline (MapGPT). This efficiency stems from our optimized prompt design, which selectively distills visual information rather than providing the redundant inputs used in MapGPT. Regarding completion tokens, NavGPT’s multi-agent communication leads to a significantly higher output volume, whereas DACo maintains a completion overhead comparable to that of MapGPT.

Overall, although DACo is not the most computationally lightweight configuration, its overhead remains moderate and substantially lower than existing multi-agent approaches.
Considering the trade-off between navigation accuracy and operational cost, DACo offers a favorable balance suitable for practical VLN deployment.

\begin{figure}[!t]
    \centering
    \includegraphics[width=0.8\textwidth]{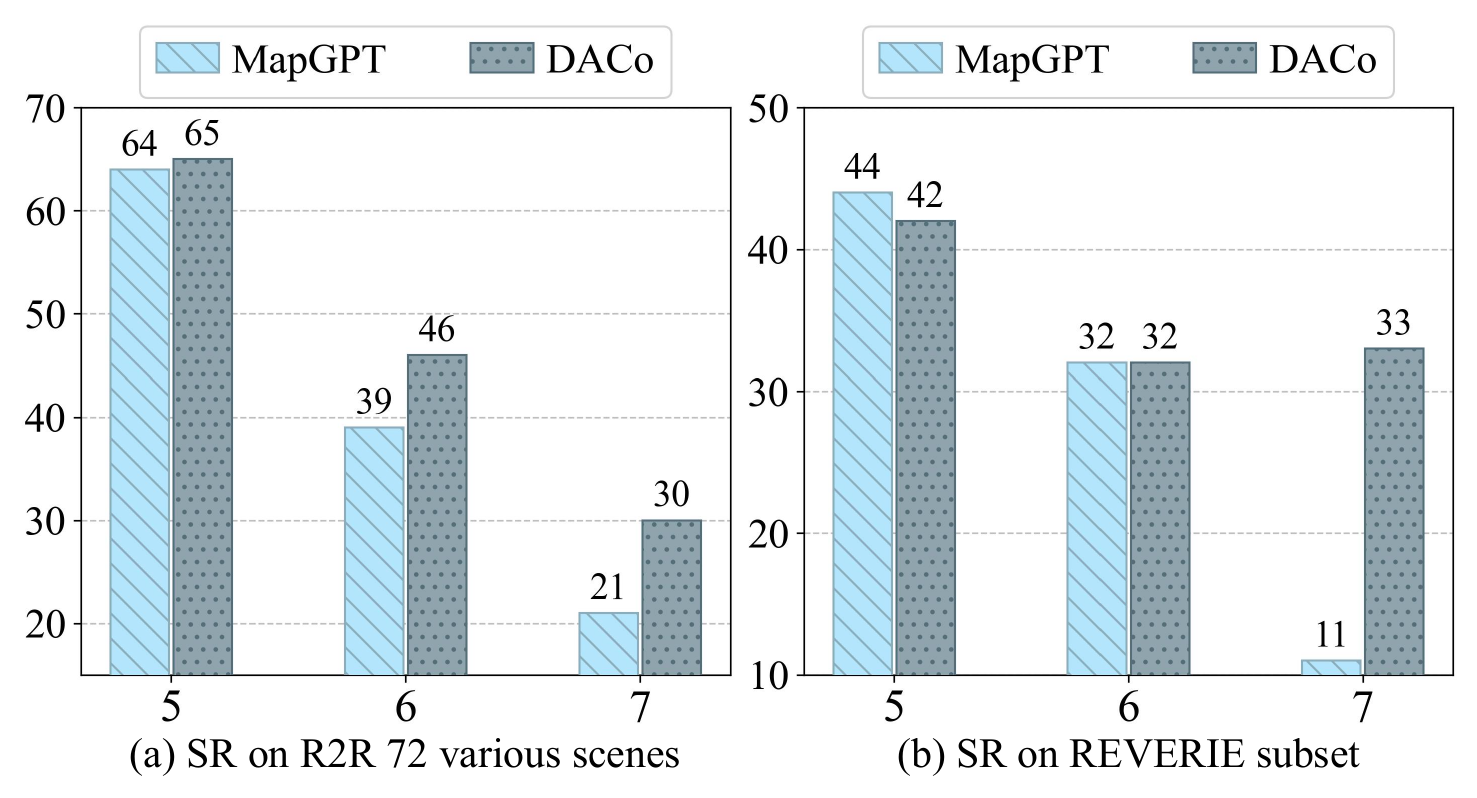}
    \vspace{-2mm}
   \caption{Results corresponding to samples of different steps. MapGPT is reproduced with GPT-4o.}
    \label{fig:steps-results}
\end{figure}

\section{Discussion}

In this section, we further analyze the capabilities of DACo and summarize key findings from our experiments.
We also provide qualitative examples to illustrate its operational workflow and examine its behavior in complex navigation scenarios.

\subsection{Analysis on Different Navigation Distance}
\label{sec:distance-analysis}
We also compared the performance of our approach with the primary baseline, MapGPT, on instructions of different lengths, as shown in Fig.~\ref{fig:steps-results}. In both datasets, an instruction typically takes 5 to 7 steps to complete. The more steps required, the greater the difficulty and the lower the SR. As shown in Fig.~\ref{fig:steps-results}(a), for samples requiring 5 steps, DACo's SR is roughly on par with MapGPT's.
However, for samples of length 7, DACo surpasses MapGPT by 9\%.
A similar trend is observed in REVERIE (Fig.~\ref{fig:steps-results}(b)): for shorter samples, DACo lags slightly behind MapGPT, but for samples of length 7, DACo's success rate leads significantly by up to over 20\%. This demonstrates that our method, the dual-agent framework, exhibits superior robustness when handling long-horizon navigation tasks.

\begin{table}[!t]
\centering
\caption{Analysis of stochasticity. All experiments are conducted on the REVERIE subset using Qwen3-VL-8B.}
\fontsize{9}{10}\selectfont
\setlength{\tabcolsep}{3.0mm}
\begin{tabular}{l|llllllll}
\toprule
\multirow{2}{*}{\textbf{Method}} & \multicolumn{4}{c}{\textbf{SR}} & \multicolumn{4}{c}{\textbf{OSR}}\\
\cmidrule(lr){2-5} \cmidrule(l){6-9}
& \textbf{Mean$\uparrow$} & \textbf{Range$\downarrow$} & \textbf{SD$\downarrow$} & \textbf{CV$\downarrow$} & \textbf{Mean$\uparrow$} & \textbf{Range$\downarrow$} & \textbf{SD$\downarrow$} & \textbf{CV$\downarrow$}\\
\midrule
NavGPT & 25.5 & 4.00 & 2.11 & 8.22\% & 35.83 & \textbf{4.00} & \textbf{2.08} & 5.81\% \\
MapGPT & 30.50 & 5.00 & 2.65 & 8.67\% & 43.83 & 6.00 &  3.33 & 7.60\% \\
DACo & \textbf{32.50} & \textbf{3.00} & \textbf{1.50} & \textbf{4.62\%} & \textbf{49.00} & 5.00 & 2.65 & \textbf{5.40\%} \\
\bottomrule
\end{tabular}
\label{tab:stochasticity-results}
\end{table}
\subsection{Analysis of Performance Variability}

A common challenge in LLM/LVLM-based methods is the inherent stochasticity of model outputs. Despite setting the temperature to zero, we still observed fluctuations during evaluation. To quantitatively assess this, we report the mean, range, standard deviation (SD), and coefficient of variation (CV) for the SR and OSR metrics over three runs, as shown in Table~\ref{tab:stochasticity-results}. CV, defined as the ratio of the standard deviation to the mean, serves as a normalized measure of dispersion to facilitate comparisons across different performance levels.
For the SR metric, DACo achieves the highest mean while exhibiting the lowest range, SD, and CV. For OSR, although DACo does not attain the lowest absolute range or SD, it yields the minimum CV, indicating comparatively stable performance when accounting for scale differences. 
These results suggest that DACo maintains strong consistency relative to existing baselines.
We hypothesize that this stability stems from the architectural separation of global planning and local execution, which reduces reasoning overload in monolithic designs and limits error propagation in heavily coordinated multi-agent systems.

\subsection{Qualitative Analysis}
\label{sec:case_study}
In this section, we present four representative cases on the R4R or REVERIE dataset to illustrate how DACo handles complex navigation scenarios through dual-agent collaboration.

\begin{figure}[htbp]
    \centering
    \includegraphics[width=0.85\textwidth]{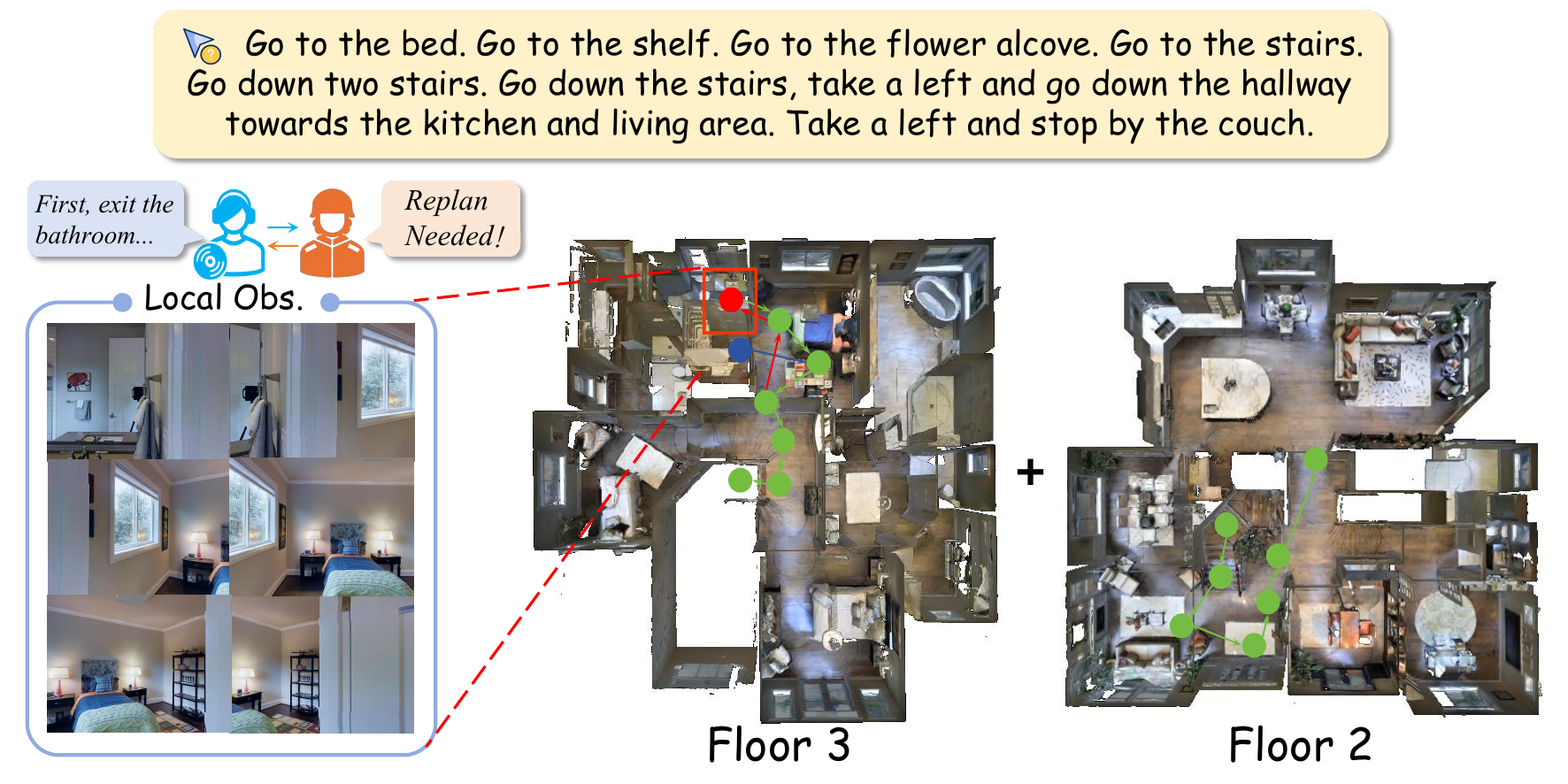}
    \vspace{-2mm}
   \caption{A case demonstrating the effectiveness of the replanning mechanism. Following a replan, DACo successfully recovers from an off-track trajectory. Blue indicates the historical trajectory, red represents the deviated trajectory, and green denotes the corrected, target trajectory.}
    \label{fig:case_study_replan}
\end{figure}

\textbf{Case 1: Target Re-planning.}
As shown in Fig.~\ref{fig:case_study_replan}, the first case highlights DACo’s ability to dynamically revise plans during execution. The instruction requires multiple intermediate stops across floors, including directional changes and sub-goals.
Initially, the Local Agent executes the navigation by following the global plan and instructions. As illustrated, the agent wanders around the bed area but fails to transition to the subsequent phase—`go to the shelf'—likely due to the sub-optimal quality of the initial global plan. However, upon entering the bathroom, the Local Agent detects a semantic discrepancy between the current local observations and the instruction, identifying that the plan is no longer viable. Consequently, the Local Agent issues a re-planning request. Based on the agent's current coordinates, the Global Agent generates an updated plan, instructing the Local Agent to first exit the bathroom. Following this revised guidance, the Local Agent not only successfully vacates the area but also identifies the corridor and stairs, ultimately reaching the goal. 
This case demonstrates how our dual-agent system enhances navigational robustness through continuous mutual verification and error correction, ensuring highly reliable performance.

\begin{figure}[t]
    \centering
    \includegraphics[width=0.85\textwidth]{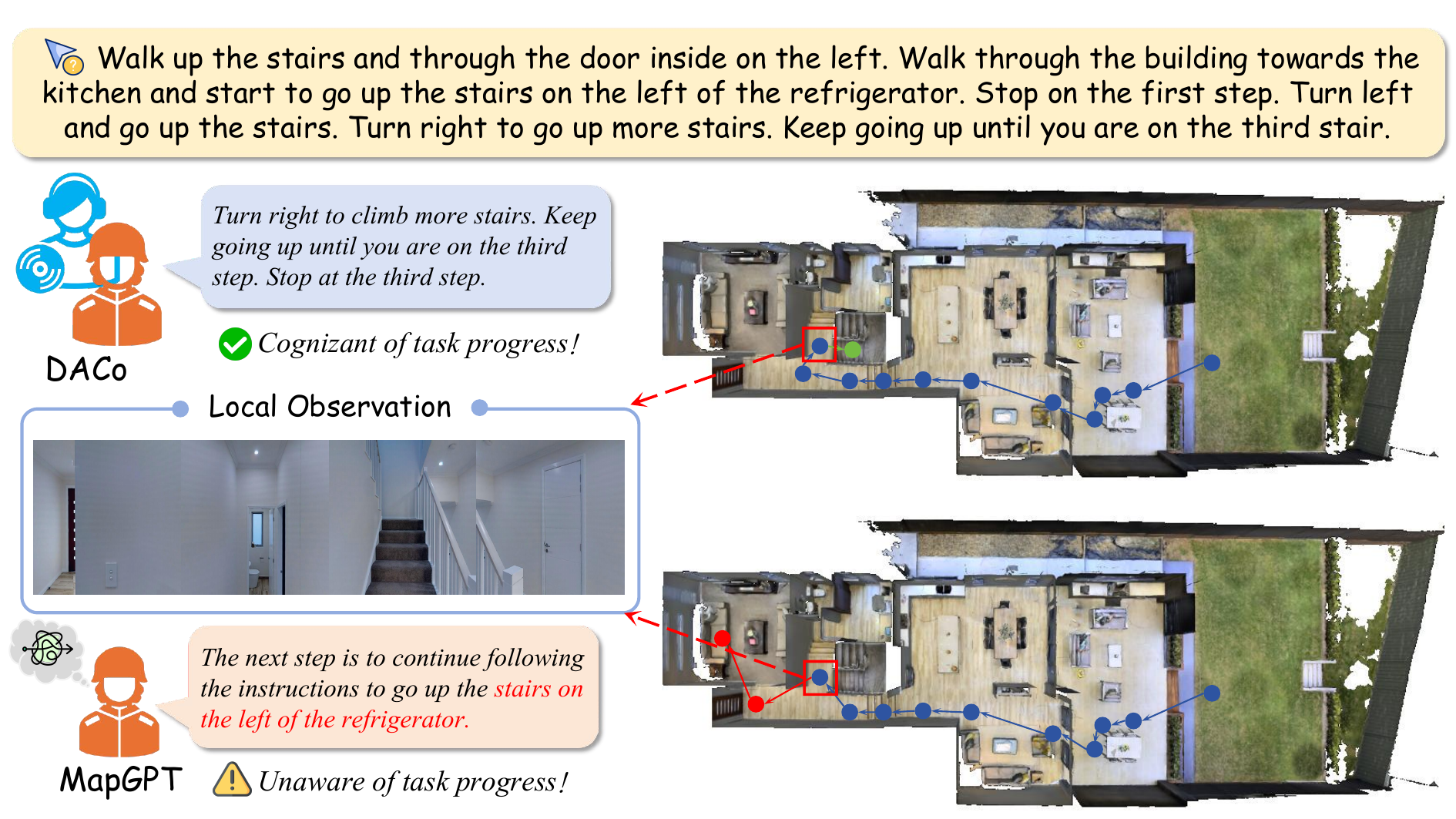}
    \vspace{-2mm}
   \caption{A comparison case between baseline and our method showcasing the long-horizon navigation capability. While baseline often suffers from orientation loss and trajectory drift, our approach maintains higher coherence and robustness. Blue indicates the historical trajectory, red represents the deviated trajectory, and green denotes the corrected, target trajectory.}
    \label{fig:case_study_long}
\end{figure}
\textbf{Case 2: Long-horizon Planning.}
As illustrated in Fig.~\ref{fig:case_study_long}, this case demonstrates a common challenge in zero-shot VLN: long-horizon instruction following. 
As previously discussed, navigation trajectories in R4R are substantially longer than those in R2R or REVERIE, thereby imposing significantly greater demands on long-range navigation stability and sustained reasoning capability.
In such scenarios, MapGPT exhibits a ``trajectory deviation" phenomenon. Toward the end of the navigation, after traversing more than twenty viewpoints, MapGPT encounters an excessively large historical context. This manifests intuitively as a loss of task progression awareness: the agent fails to track completed sub-tasks and erroneously designates already finished segments as current objectives (as highlighted in red in Fig.~\ref{fig:case_study_long}). We attribute this failure to the inherent structural limitations of the single-agent paradigm, where the lack of hierarchical task management leads to cognitive drift under prolonged environmental exposure.
In contrast, our system incorporates a dynamic planning mechanism, enabling the Global Agent to maintain real-time supervision of the Local Agent from a holistic perspective.
In Fig.~\ref{fig:case_study_long}, the global plan (highlighted in the blue dialog box) is adaptively formulated based on the Local Agent’s current coordinates and historical trajectory. This strategic guidance prevents the Local Agent from committing the same errors observed in MapGPT. Furthermore, our prompt design is significantly more streamlined than that of MapGPT, as it selectively omits redundant visual information. These mechanisms collectively enable DACo to maintain precise alignment between task progression and original instructions even in long-horizon scenarios, effectively mitigating the risk of disorientation.

\begin{figure}[htbp]
    \centering
    \includegraphics[width=0.85\textwidth]{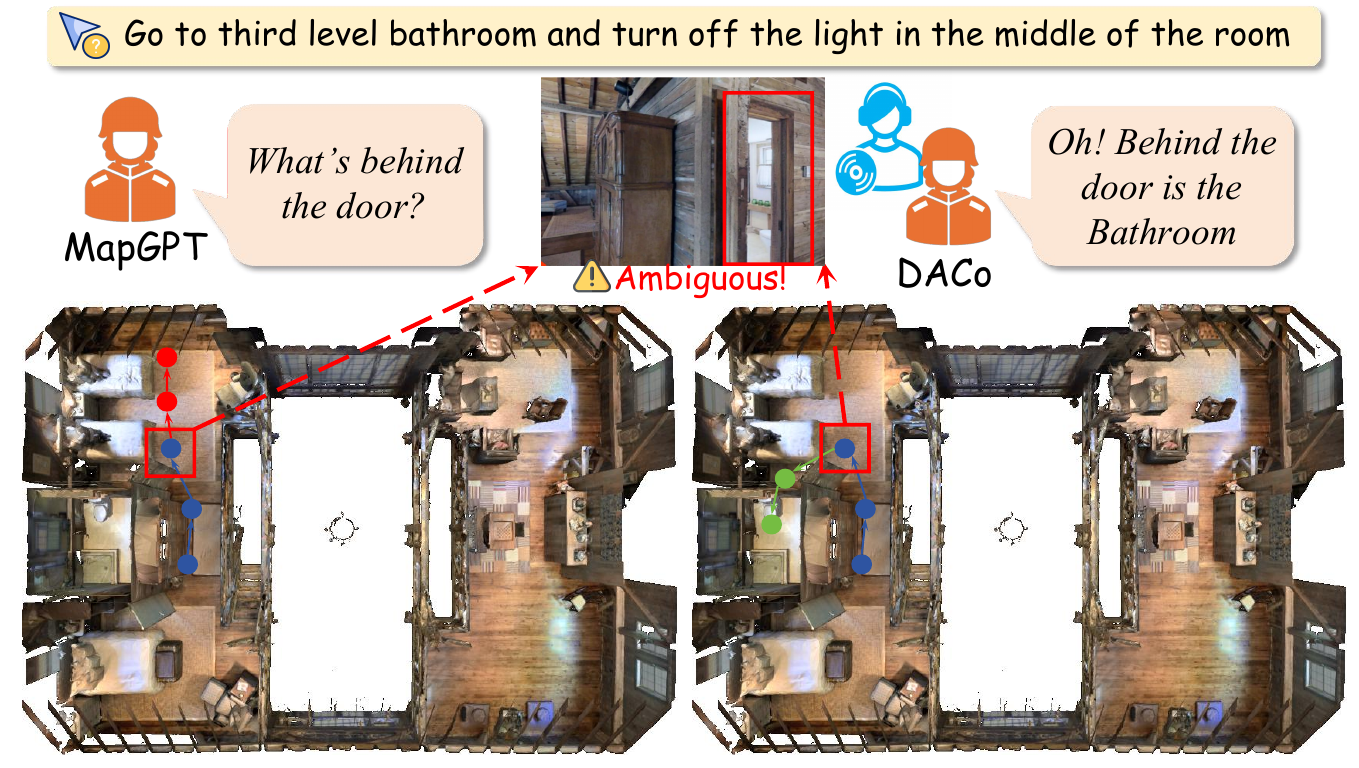}
    \vspace{-2mm}
   \caption{A comparison case between baseline and our method showcasing the strong spatial layout understanding.}
    \label{fig:more_cases_1}
\end{figure}

\textbf{Case 3: Spatial Layout Understanding.} In this case, we present another example to demonstrate that our method can address another major challenge in the VLN domain: how to navigate when the target is occluded. Previous VLN methods only receive egocentric information, which may prevent the agent from exploring the right viewpoint when the target landmark is occluded. As shown in Fig~\ref{fig:more_cases_1}, the instruction in this case is ``Go to third level bathroom and turn off the light in the middle of the room". We still focus on the comparison of the trajectories generated by MapGPT and DACo. Step 3 marks the point where the two methods diverge; the key local observations perceived by the agent at this juncture are also visualized in Fig.~\ref{fig:more_cases_1}. Humans can easily recognize that there might be a bathroom behind the door, as the toilet is partially visible. However, the occlusion caused by the wall significantly affects the LVLM’s judgment. If only egocentric information is available, the agent would not know to enter through this door, leading it to deviate from the ground truth trajectory. Benefiting from the Global Agent, our Local Agent receives prior knowledge indicating that the room behind the door is indeed a bathroom, enabling it to complete the navigation correctly. This case highlights DACo's robust proficiency in spatial layout understanding, enabling it to navigate successfully even in scenarios characterized by severe visual occlusions.

\begin{figure}[t]
    \centering
    \includegraphics[width=0.85\textwidth]{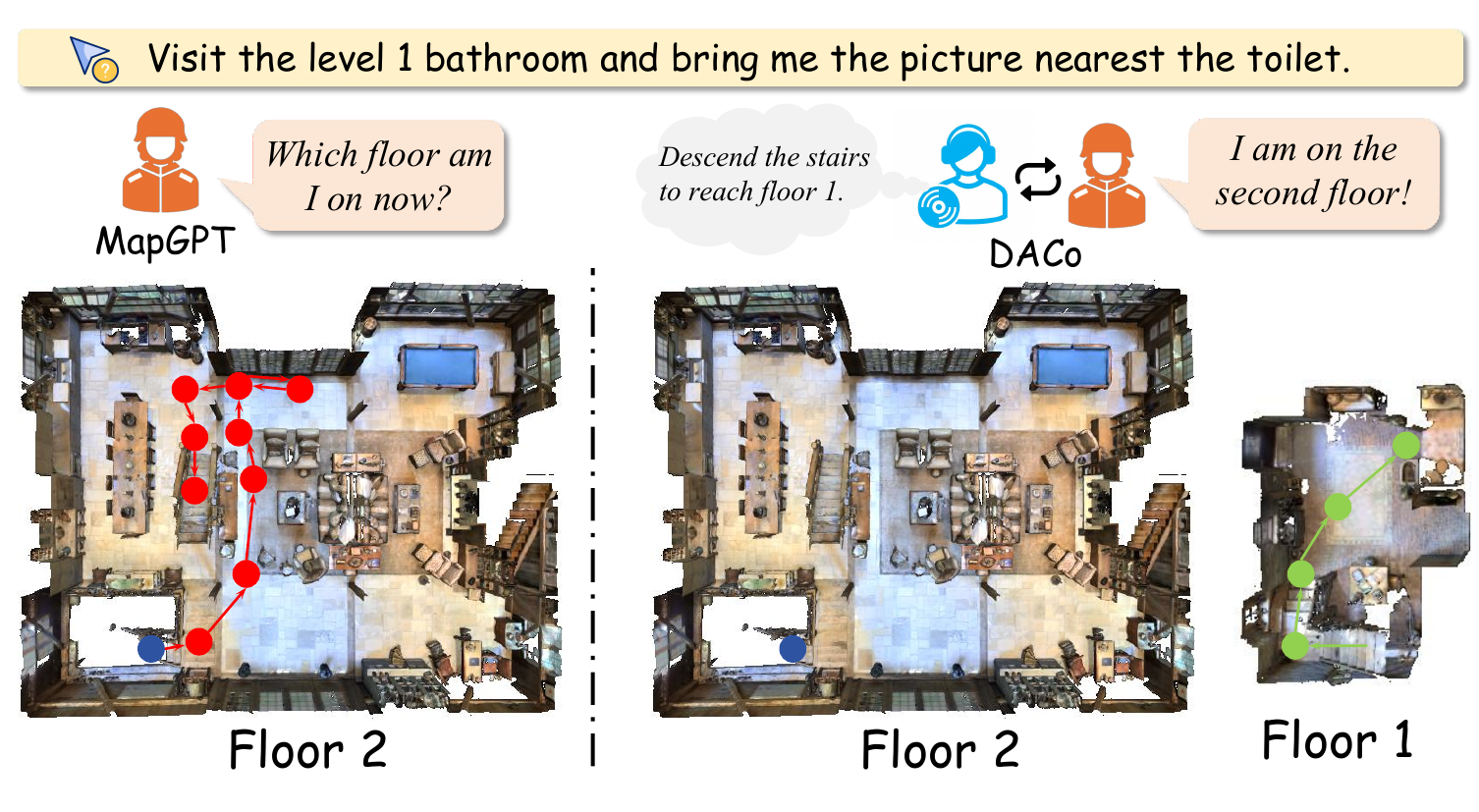}
    \vspace{-2mm}
   \caption{A comparison case between baseline and our method showcasing the multi-level navigation capability.}
    \label{fig:more_cases_2}
\end{figure}

\textbf{Case 4: Cross-level Navigation.} 
In Fig.~\ref{fig:more_cases_2}, we present another successful case of DACo on REVERIE, illustrating how the DACo framework accomplishes a complex navigation task in a multi-level building through dual-agent collaboration. This case also illustrates a common challenge in VLN: the difficulty current zeroshot agents face in cross-floor navigation. Tasks involving multiple floors are prevalent both in real-world scenarios and in standard benchmarks. Specifically, 70\% scenes in the REVERIE validation unseen set are multi-level buildings. However, the agent initially lacks knowledge of its current level, and REVERIE instructions do not provide explicit step-by-step guidance. Consequently, as shown in the left half of Fig.~\ref{fig:more_cases_2}, without external planning support, MapGPT fails to descend to level 1 and remains wandering aimlessly on the second floor. Benefiting from the dual-agent mechanism, the global agent generates a detailed and executable trajectory for the local agent before navigation. In this instance, the global agent correctly infers that the local agent is on level 2 and must first descend to level 1. Ultimately, DACo successfully reaches the target location, with its trajectory fully aligned with the ground truth. This case demonstrates that DACo possesses superior global spatial awareness, enabling it to handle complex, multi-level navigation tasks that require high-level planning and adaptive execution.

\textbf{Overall Discussion.}
In these cases, DACo ultimately reaches the correct target location with a trajectory that closely follows the ground-truth path. These cases demonstrate that DACo possesses stronger global spatial reasoning and adaptive planning capabilities, particularly in complex long-horizon and cross-level environments.

\section{Conclusion}
In this work, we introduce DACo, a novel dual-agent collaboration framework for zero-shot indoor scene navigation.
Technically, DACo decomposes the scene navigation task into a Global Commander for global intent-level planning and a Local Operative for local action-level execution.
To achieve closed-loop reasoning, we introduce dynamic subgoal planning and path replanning mechanisms that enable adaptive coordination between the two agents.
Experimental results show that DACo consistently outperforms existing zero-shot methods on the R2R, REVERIE, and R4R benchmarks.
Remarkably, DACo enables open-source backbones (e.g., the Qwen series) to outperform baseline methods integrated with proprietary models like GPT-4o, demonstrating its architectural superiority and robust generalization.

\textbf{Limitations.}
Despite its effectiveness, DACo has several limitations. First, it relies on Bird’s-Eye View (BEV) maps for global guidance, which can be challenging to acquire in some real-world settings. However, our framework only requires coarse-grained representations, and its design provides a viable paradigm as mapping technologies mature~\cite{chandaka2025humanlikenavigationworldbuilt}. Second, DACo inherits the stochasticity and backbone dependency common to all LVLM-based agentic frameworks. Nevertheless, given the scarcity of real-world navigation data, we believe a robust zero-shot approach is invaluable, with its generalization benefits significantly outweighing these inherent constraints. Overall, DACo offers a valuable and practical foundation for future research in Vision-Language Navigation.

\textbf{Future Work.}
While DACo establishes a novel paradigm for indoor VLN, several avenues remain for future exploration. First, we aim to transition our framework into continuous environments to bridge the gap between simulation and real-world deployment. Second, we plan to investigate more sophisticated spatial modeling techniques for global map construction, thereby reducing the current reliance on pre-processed top-down scene views. Finally, we intend to extend our methodology to outdoor navigation scenarios, further bolstering the generalization and robustness of the system across diverse environmental scales.

\bibliographystyle{unsrtnat}
\bibliography{references}


\appendix

\section{Implementation Details}
\label{implementation_details}


\subsection{Prompt Details}
\label{prompt_details}
In this section, we will introduce key prompts designed for DACo. In the action prompt of local agent part, our prompt design draws on insights from prior work\cite{mapgpt} while incorporating adaptations tailored to DACo's framework. Due to space limitations, we omit some parts that are redundant with prior work.

\begin{tcolorbox}[colback=blue!5!white,colframe=blue!75!black,title=Action Prompt of Local Agent, breakable]
You are a first-person mobile robot navigating inside a house to complete a vision-language navigation (VLN) task. You are collaborating with a partner who has access to the global Top-down View image.

\medskip
At \textbf{each step}, you are provided with the information below:

1. \textbf{Instruction} is ...

2. \textbf{Global Plan} is a step-by-step path that your partner told you, which is much more detailed and important for you to refer.

3. \textbf{History} is ...

4. \textbf{Map}: is ...

5. \textbf{Action options} are ...

\medskip
\textbf{Reasoning steps:}
\begin{enumerate}
  \item First, you need to align \textbf{Global Plan} and \textbf{Instruction} with \textbf{History} to estimate your instruction execution progress.  
  \item Second, for each \textbf{Action Option}, you should combine the \textbf{Instruction} and \textbf{Global Plan}, carefully examining the relevant information, such as scene descriptions, landmarks, and objects.   
  \item Before you give the final decision, check the Place IDs in the \textbf{History}, avoiding repeated exploration that leads to getting stuck in a loop.   
  \item ...  
  \item If you are very sure that you have arrived at the destination, you can choose the ``Stop" action.
\end{enumerate}

\medskip
\textbf{Additional Tips:}
\begin{enumerate}
    \item When combining the text with the images, you must carefully consider the relationship between the verbs and landmarks in the `Global Plan' and the objects in the images. 
    \item If you find that the current scene is impossible to carry out the instruction, consider backtracking to a previously visited location.
    \item If the global plan seems incorrect or lacks detail, feel free to trigger a `\textbf{replan}' so your partner can refine the objectives.
\end{enumerate}

\medskip
\textbf{Output format:}  

1. Your answer should include two parts: ``Thought", and ``Action".   

2. \textbf{``Thought"}: ... 

3. \textbf{``Action"}: ...

\medskip
\textbf{Example:}

Thought: something you output about your thought or reasoning process.

Action: A

\end{tcolorbox}

\begin{tcolorbox}[colback=blue!5!white,colframe=blue!75!black,title=Planning Prompt of Global Agent, breakable]
You are collaborating with a robot agent to finish the VLN task. 
Your task is to plan a global and detailed path based on a given Top-down View image and a navigation instruction to help your partner. 

\medskip
At \textbf{each step}, you are provided with the information below:

1. \textbf{Instruction} is a global guidance, but you might have already executed some of the commands. You need to carefully discern the commands that have not been executed.

2. \textbf{Current Observation} is text that describe your current observation.

3. \textbf{Previous Plan} records previous long-term multi-step plan info that made by you in the last step.

3. \textbf{Top-down View Images} is a list composed of Top-down View images, where each image corresponds to the top-down view of one floor of a building. The agent's trajectory is annotated on the corresponding Top-down View map, red denotes the starting point, blue denotes the position reached at the `i'th step, green denotes the current point.

\medskip
\textbf{Reasoning steps:}
\begin{enumerate}
  \item First, try to find the green circle marked with ``now" on \textbf{Top-down View Images} and you can figure out the agent's current floor.
  \item Secondly, you need to infer that the agent's execution progress based on the \textbf{Instruction} and trajectory points marked on \textbf{Top-down Images}.
  \item Thirdly, you need to determine the agent's initial orientation based on the \textbf{Current Observation} and trajectory points in \textbf{Top-down View Images}. 
  \item Then, you need to \textbf{plan a new detailed global path}. If you think the agent is close enough to the destination, you can suggest him to stop.   
\end{enumerate}

\medskip
\textbf{Output format:}

1. Your answer should be JSON format and must include two fields: ``Thought" and ``New Plan". 

2. \textbf{``Thought"}: Think about what to do and why, and complete your thinking into 
``Thought".

3. ``New Plan": You also need to update your new multi-step path plan to ``New Plan".

\medskip
\textbf{Example:}

\{
    ``Thought": ``...",
    ``New Plan": ``..."
\}

\end{tcolorbox}

For REVERIE dataset, because we only focus on navigation part in instruction, so the ``Instruction" description is attached with the following content: \textit{``(Instruction is a high-level command that directs an agent to complete a certain task at a specific location. It is worth noting that the agent does not need to complete the task itself; you only need to provide a navigation path to the target location.)"}

\begin{tcolorbox}[colback=blue!5!white,colframe=blue!75!black,title=Replan Prompt of Global Agent, breakable]
You are a 3D-aware embodied navigation agent. Your task is to replan a navigation path in an indoor environment.

\medskip
\textbf{Input: (same with Planning Prompt)}
\begin{enumerate}
  \item Goal Location: A high-level goal description (may not include full details).
  \item Top-down View Images: ...
  \item Current Observation: ...
\end{enumerate}

\medskip
\textbf{Task}

Your task is to analyze the Global Instruction and the Top-down View images, then replan a valid and step-by-step navigation path from the Current Location (red) to the final target location specified by the instruction.

\medskip
\textbf{Planning Rules}
\begin{enumerate}
  \item Infer the goal location based on the start location (red) and instruction
  \item Always localize your position (floor, landmark, heading if known) based on the green mark.
  \item Ensure the new path connects logically to the current point (red) and the global goal.
  \item If stairs are needed, identify them clearly and describe floor transitions.
  \item Avoid walls or void regions in the Top-down View map.
  \item End the plan with a stop action at the final target.
\end{enumerate}

\textbf{Action Language (strict)}
\begin{enumerate}
    \item Each step is one verb from this closed set: move forward, ...
    \item Grammar: $\langle$verb$\rangle$ + $\langle$target landmark$\rangle$
    \item Every move forward must include an until/to/past clause (distance or clear landmark).
    \item Every floor change must name the connector and the destination floor (e.g., to Floor 2).
    \item The final step must be stop with the final landmark.
\end{enumerate}

\textbf{Output Format}

Return a raw JSON object with exactly two fields:
\begin{enumerate}
    \item ``Thought": Reasoning process combining Global Instruction, Current Location (blue), Start Location (red), and BEV images. State any assumptions or missing info.
    \item ``New Plan": Based on your `BEV Images' and current `Thought', you also need to provide your new multi-step path plan to `New Plan'.
\end{enumerate}

\end{tcolorbox}

\begin{tcolorbox}[colback=blue!5!white,colframe=blue!75!black,title=Location Description Prompt , breakable]
You are collaborating to do a VLN task in a house with a partner who is only access to the global Top-down View image. Your task is to describe your current location based on a sequence of images.     

\medskip
Each image is accompanied by a caption, such as ``Scene 1 (on your right)". Please summarize the environment in front of, to the left of, to the right of, and behind you. Finally, if possible, you can infer which room you are currently in.

\medskip
\textbf{Output Format:}
Please summarize your output into one paragraph of \textbf{plain text}, and the output shouldn't be too long. Do not output markdown format!
\end{tcolorbox}

\end{document}